\documentclass[journal]{IEEEtran}
\IEEEoverridecommandlockouts
\usepackage{cite}
\usepackage{amsmath,amssymb,amsfonts}
\usepackage{algorithm}
\usepackage{algorithmicx}
\usepackage{graphicx}
\usepackage{textcomp}
\usepackage{tabu}
\usepackage{xcolor}
\usepackage[justification=centering]{caption}
\usepackage{algpseudocode	}
\makeatletter
\newcommand{\removelatexerror}{\let\@latex@error\@gobble}
\makeatother
\usepackage{comment}
\usepackage{booktabs}
\usepackage{soul}
\usepackage[pdftex,bookmarks=true]{hyperref}

\soulregister\cite7 
\soulregister\citep7 
\soulregister\citet7 
\soulregister\ref7
\soulregister\pageref7

\def\BibTeX{{\rm B\kern-.05em{\sc i\kern-.025em b}\kern-.08em
    T\kern-.1667em\lower.7ex\hbox{E}\kern-.125emX}}

\begin{document}

\title{AI Based Digital Twin Model for Cattle Caring}

\author{
\begin{tabular}{c}
Xue Han, Zihuai Lin, Cameron Clark, Branka Vucetic\tabularnewline
\tabularnewline
\end{tabular}
}



\maketitle

\begin{abstract}
In this paper, we developed innovative digital twins of cattle status that are powered by artificial intelligence (AI). The work was built on a farm IoT system that remotely monitors and tracks the state of cattle. A digital twin model of cattle health based on Deep Learning (DL) was generated using the sensor data acquired from the farm IoT system. The health and physiological cycle of cattle can be monitored in real time, and the state of the next physiological cycle of cattle can be anticipated using this model. The basis of this work is the vast amount of data which is required to validate the legitimacy of the digital twins model. In terms of behavioural state, it was found that the cattle treated with a combination of topical anaesthetic and meloxicam exhibits the least pain reaction. The digital twins model developed in this work can be used to monitor the health of cattle.
\end{abstract}




\maketitle
\section{Introduction}\label{sec:introduction}
\IEEEPARstart{D}{igital} twins are virtual digital representations of physical objects, in which the physical object and its corresponding virtual digital representation interacts remotely in real time\cite{1}. A digital twin model incorporates multi-disciplinary, multi-physical quantity, multi-scale, and multi-probability simulation processes and fully utilises physical models, sensor updates, operation histories, and other data\cite{2}. In addition, digital twins complete the mapping in virtual space so that the full life cycle process of associated entity equipment is reflected\cite{3}. Digital twins are a transcendental idea that can be regarded as one or more crucial and interdependent digital mapping systems for the actual object\cite{4,5}.

Connectivity, modularity, and autonomy between virtual and actual items can all be realised with digital twins. It can be accomplished across the whole production process from product design through product system engineering to production planning, implementation and intelligence, resulting in a self-optimizing closed loop\cite{6}. To put it another way, by connecting the actual object with the virtual number, the real object may offer the real information to optimise the digital model, and the digital model can foresee potential situations to alter the real object. The two complement each other to create a self-closing optimisation mechanism\cite{7}. Nowadays, digital twins have been increasingly employed in a variety of industries, including product design, product manufacturing, medical analysis, engineering construction and other areas\cite{8}. As a result, digital twins can be seen as a major force behind the intelligent manufacturing paradigm\cite{9}. Digital twins have recently been deployed in a variety of fields, including livestock farming\cite{10,11}. In today's animal husbandry, testing the health of cattle is a crucial stage. Large cow ranches, in particular, need to keep track on their herd's health in real time\cite{12,13}. 

Deep learning (DL) is a new direction in machine learning that is being introduced to bring it closer to the goal of AI and has made tremendous progress in solving issues that were previously unsolvable in AI \cite{14,15,16,17,18,19}. It has proven to be so effective in detecting complicated structures in high-dimensional data that it might be used in a wide range of scientific, business and government applications\cite{20,21,22}. The long short-term memory network (LSTM) is a type of cyclic neural network and one of the deep learning algorithms that can analyse and forecast critical time with very long intervals and delays in time series\cite{23,24}. 

In a long time series, the LSTM neural network algorithm can determine which information should be stored and which should be discarded\cite{25}. The development of digital twins relies heavily on accurate time series prediction. Internal and external disruptions might result in time series that are exceedingly nonlinear and random. Complex objects time series prediction may be employed at any stage of their life cycle, which is also a major component of the digital twin model\cite{26,27}. Therefore, it is extremely dependable to use LSTM model to build digital twins. 

The goal of this study is to develop an intelligent digital twins technique with LSTM neural network to provide a variety of behavioural detection and prediction of cattle's health status, including current health analysis and upcoming physiological cycle, etc. The digital twins concept is heavily reliant on massive volumes of data reflecting cattle location, movement and free grazing time, etc, collected by IoT monitoring systems, e.g., \cite{28,29,30},  wireless sensor networks, e.g., \cite{31,32,33,34,35,36} or cellular systems \cite{37,38,39,40}.

The outline of the paper is given below. First in \autoref{sec:related}, existing works are described and reviewed for understanding of relationship between behavioral state and pain in cattle. After that, in \autoref{system}, the new digital twin system involved in this study is illustrated and explained in detail. In \autoref{sec:Datamining}, necessary data mining and data analysing for the IoT system are carried out. In this part, most of the data processing work are accomplished with the help of MATLAB. In \autoref{mainly}, cattle's behaviour states are modeled by training the LSTM neural network in the digital twin model and cattle's states in the next cycle is predicted by using this deep learning technique. In \autoref{Result}, The accuracy of the trained LSTM model is discussed and verified, the best pain treatment for the cattle could be determined. Finally, \autoref{sec:Conclusion} deduces a proper conclusion.

\section{Related Work}\label{sec:related}
Analysing the relationship between cattle's states and their level of pain is a crucial step to obtain the cattle's health. Because animals' assessments of changes in pain behaviour are highly subjective, it is difficult to apply a single index to assess and judge the severity of pain for cattle. Several studies have evaluated cortisol responses in cattle to establish the intensity and duration of pain related to castration. Cortisol has long been employed as a pain indicator since the magnitude of its reaction, such as peak height, duration of response, and combined response, frequently correlates with the expected toxicity of specific operations. The cattle who had surgery (i.e., those in more pain) ate grass less frequently, took fewer steps with pedometers, spent more time standing, and had symptoms of aberrant behaviour than those in the control group\cite{41,42}. Another study looked at the association between cow pain levels and various state postures\cite{43,44,45}. It stated that the cattle's pain level is affected by their attention, head position, ear position, face expression, and back position and that the pain level is split into three stages. The cattle that are passive without meals, grooming, or sleeping, for example, have pain on a scale of 1. In contrast, the cattle that are accompanied by lying down immediately after coming out of bed have pain on a scale of 2. if the cattle is active and attention towards the surroundings, it can be regarded as pain on a scale of 0.

Other research backs up these claims, claiming that pain is a subjective emotional state that is difficult to quantify objectively\cite{46}. However, similar pain behaviours, which can be defined as those that occur in the presence or absence of pain, can be used to assess discomfort. After calves are burned or dehorned, postoperative physical pain is associated with increased head-related motor behaviours, such as head shaking and ear tossing. Specific postural alterations, such as limb protection, are also linked to pain disorders. A pedometer that records the number of steps taken has been employed as a significant indicator of pain behaviour monitoring. Cattle in pain take fewer steps, according to some research, and the number of steps taken may help assess behavioural changes following painful manipulations\cite{47}. Cattle eating behaviour and intake can be monitored to detect potential changes in health or pain state, with reduced intake in cattle in pain\cite{47}.

At present, there is little previous work using the digital twin to analyze the behavior of cattle in order to monitor their health, because there is little work using the digital twin in animal husbandry. In addition, compared to previous work, e.g., \cite{3,7,8,10}, we use a large amount of data collected from the farm Internet of Things to analyze, model, and support our results.

\section{The Proposed Digital Twin Model}\label{system}
\begin{figure}[ht]
    \centering
    \includegraphics[width=0.45\textwidth]{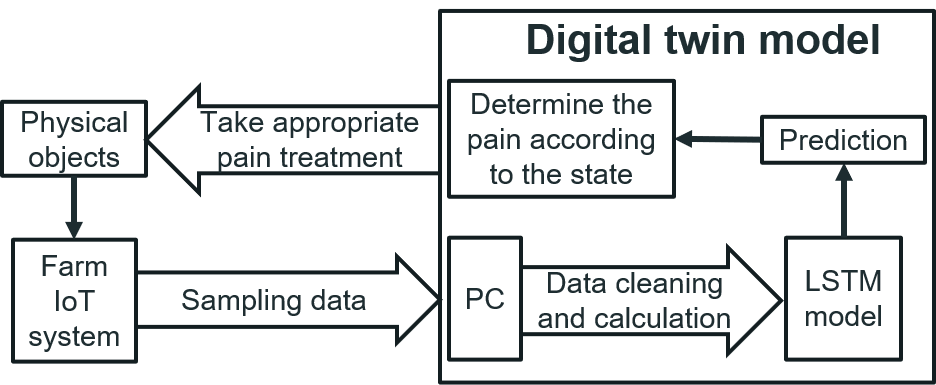}
    \caption{The digital twin model of the cattle.}
    \label{digitaltwinmodel}
\end{figure}

In this work, we propose a digital twin model for cow health status monitoring, animal well-being prediction and appropriate pain treatment. Digital twins are virtual digital models that combine with real-world things. In our proposed model, the farm IoT system first collects relevant state data from physical items and transmits it to a cloud server. The prediction of state dynamics is then completed using the LSTM model of cow state following a data processing sequence and calculations. Actual data and expected results are used to determine the pain and health status of cattle. Finally, the appropriate pain therapy is implemented based on the cow's projected discomfort status. Simultaneously, new sample data is entered into the cloud server and compared to the prior model's anticipated value to continuously modify and optimize the model. This method completes the interaction between the virtual digital model and the real-world physical object. The block diagram of the digital twin model is depicted by \autoref{digitaltwinmodel}. 

In the following, we will describe the model step by step. We will start with the process of data mining and analysing, follow by the development of an LSTM model for cattle health status prediction, then determine the pain treatment. 

\section{Data Mining and Analysing}\label{sec:Datamining}
This section primarily discusses the processing method of the data sets, i.e., the original data measured by sensors of the farm’s IoT system. This data set is systematically treated in preparation for future use of the modelling. Particularly, the data sets of the cattle's states are analysed, and a digital twin model of the cattle is produced using these data sets. A vast amount of data may be used to evaluate the model's correctness, and the state of the cattle can then be predicted.

\subsection{Data Processing}
\begin{figure}[ht]
    \centering
    \includegraphics[width=0.45\textwidth]{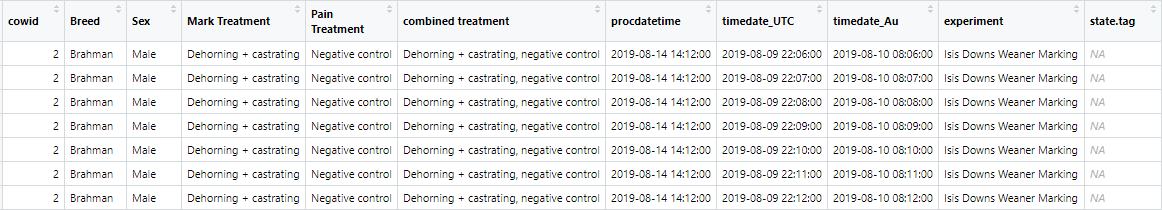}
    \caption{Part of the original data sets.}
    \label{dataset}
\end{figure}

The sensor's raw data set consists of 11 columns with over 50 million rows; a part of the original data set is displayed in \autoref{dataset}. It entails the treatment of 759 cattle of various breeds and genders in various methods. There are eight categories used to classify cattle's status: Resting, Rumination, High Activity, Medium Activity, Panting (Heavy Breathing), Grazing and Walking. Each sensor takes a minute-by-minute reading of the cows' real-time status, with each cow having 74,455 data points collected between AU\_time 8:06 am on August 10 and 1:01 am on October 1, 2019. Five cattle breeds are represented in the data sets: Angus, Brahman, Brangus, Charolais and Crossbred. There are also 13 potential therapy combinations, which are depicted in \autoref{combinedtreatment}. This section focuses on systematically processing of these data, including data segmentation, data cleaning, and data calculating.
\begin{table}[ht]
    \centering    
    \caption{The combined treatment.}
    \begin{tabular}{c}
    \hline  
    Combined treatment\\
    \hline
    Castrating, negative control(C,N)\\
    Castrating, topical anaesthetic(C,T)\\
    Castrating, meloxicam(C,M)\\
    Castrating, topical anaesthetic + meloxicam(C,T+M)\\
    Dehorning, topical anaesthetic(D,T)\\
    Dehorning, meloxicam(D,M)\\
    Dehorning, topical anaesthetic + meloxicam(D,T+M)\\
    Dehorning, negative control(D,N)\\
    Dehorning + castrating, topical anaesthetic(D+C,T)\\
    Dehorning + castrating, meloxicam(D+C,M)\\
    Dehorning + castrating, topical anaesthetic + meloxicam(D+C,T+M)\\
    Dehorning + castrating, negative control(D+C,N)\\
    Positive control(P)\\
    \hline 
    \end{tabular}
    \label{combinedtreatment}
\end{table}

\subsubsection{Data segmentation}
The first step in data processing is the segmentation. The data are grouped by cattle of the same sex, breed, and treatment. Because the original data are massive, we segment the data using RStudio and R programming language. \autoref{numberofcattle} shows the number of cows segmented and integrated; the first column of the table reflects the various abbreviations for combined treatment. As indicated in \autoref{numberofcattle}, F and M denote Female and Male, respectively and the number represents the number of animals. Cows are just dehorned, but bulls are typically castrated or castrated plus dehorned, as shown in \autoref{numberofcattle}. 

Furthermore, the number of other breeds is limited, and data is sparse except for the Brahman breed. The Brahman breed covers all of the combined treatments and is relatively numerous. This characteristic facilitates subsequent data analysis and processing. This is because a vast amount of data facilitates the analysis of overall data characteristics and avoid errors caused by individual and particular data. As a result, the resting state of Brahman's Female with Positive Control is used to demonstrate data processing and prediction.
\begin{table}[ht]
    \centering
    \caption{The number of cattle given the combined treatment.}
    \begin{tabu}{X[0.8,l]|X[0.2,l]|X[0.1,l]|X[0.35,l]|X[0.2,l]|X[0.35,l]|X[0.1,l]|X[0.4,l]|X[0.1,l]|X[0.35,l]|X[0.1,l]}
         \hline  
         Combined treatment&An-gus F&M&Brah-man F&M&Bran-gus F&M&Char-olais F&M&cross-bred F&M\\\hline
         C,M&0&1&0&1&0&1&0&0&0&1\\\hline
         C,N&0&1&0&1&0&0&0&0&0&1\\\hline
         C,T&0&1&0&3&0&2&0&0&0&6\\\hline
         C,T+M&0&1&0&1&0&0&0&0&0&1\\\hline
         D,M&0&0&70&0&1&0&2&2&13&0\\\hline
         D,N&0&0&39&0&1&0&2&2&9&0\\\hline
         D,T&0&0&101&3&3&0&4&7&20&1\\\hline
         D,T+M&0&0&66&2&4&0&3&0&12&0\\\hline
         D+C,M&0&0&0&50&0&2&0&0&0&10\\\hline
         D+C,N&0&0&0&30&0&2&0&0&0&7\\\hline
         D+C,T&0&1&0&81&0&3&0&0&0&22\\\hline
         D+C,T+M&0&0&0&50&0&1&0&0&0&13\\\hline
         P&13&14&14&5&10&0&3&1&38&0\\\hline
         Total number&13&19&290&227&19&11&14&11&92&62\\\hline 
    \end{tabu}
    \label{numberofcattle}
\end{table}

\subsubsection{Data cleaning}
As shown in \autoref{dataset}, when the sensor detects and transmits the status of the cow, it also sends a lot of invalid data. The accuracy of the original data will be considerably influenced if using these data directly. Therefore, the initial step is to clear up the corrupted data.

Because the data returned by the sensor represents the cattle's states at a specific point in time, quantifying that state is critical for further design. In this work, the time of various states each hour in minutes is taken as the research object. Because corrupted or invalid data usually aggregate, identifying the point at which incorrect data arrives as 0 is not precise and may impact subsequent calculations. Therefore, deleting the corresponding time does not influence the total calculation when cleaning the corrupted data.
\begin{figure}[ht]
    \centering
    \includegraphics[width=0.38\textwidth]{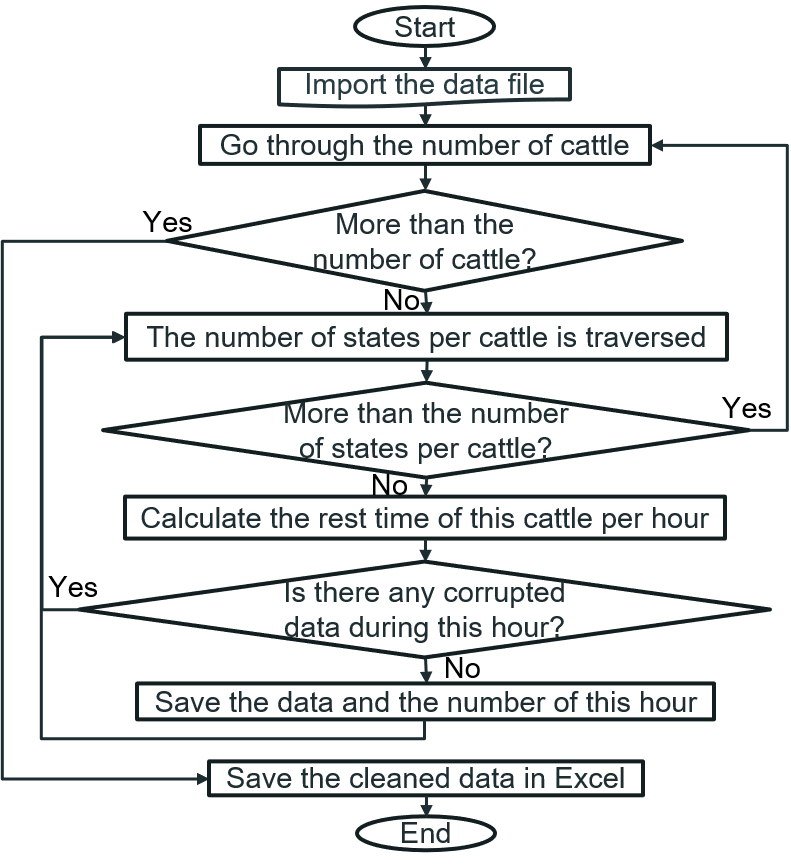}
    \caption{Process flow chart of data cleaning.}
    \label{cleandatacode}
\end{figure}

The flow chart of data cleaning is shown in \autoref{cleandatacode}. Data cleaning mainly focuses on the segmented data to clean and organize and finally obtains the cleaned data and its corresponding time series. This step primarily calculates the resting time of cattle in each hour. If it exists any corrupted data during the calculated hour, that hour's data will be destroyed.

\subsection{The state of cattle throughout the sampling period}
\begin{figure}[ht]
    \centering
    \includegraphics[width=0.38\textwidth]{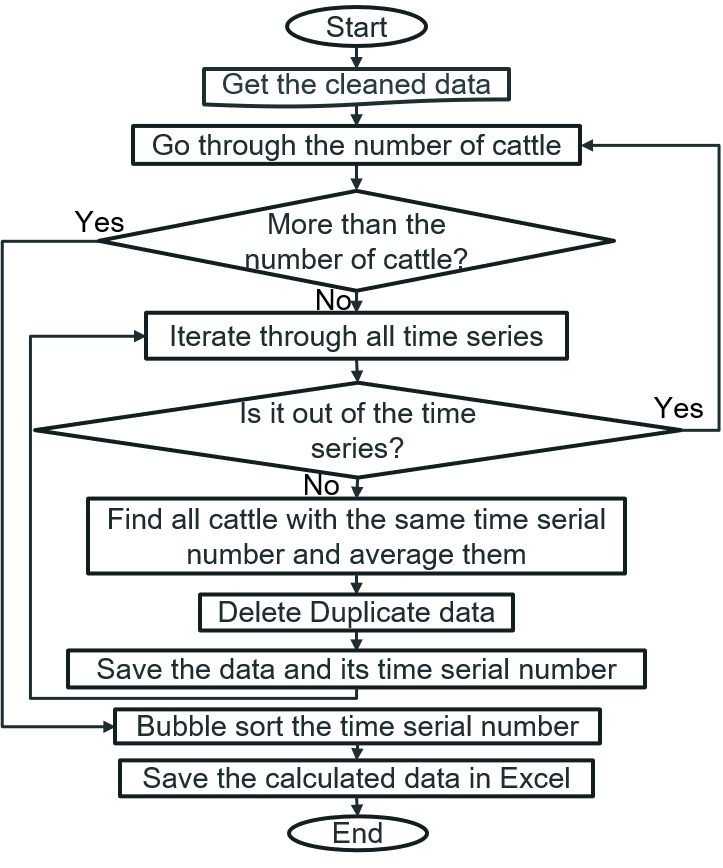}
    \caption{Process flow chart of average state times for multiple cattle.}
    \label{avecattle}
\end{figure}

Acquiring the cattle's state changes across the sample period needs to average one group's data of cows due to large amount of discrete and lost data from a single cow. For example, averaging the resting time per hour of 14 Brahman females under Positive control can determine variations in the resting state of Brahman treated with Positive control during the sample period. The time series after data cleaning are different between each cattle's data set, since invalid data collected by sensors in the farm's IoT system is usually a random process. Therefore, the data processing in this step is to average the state data of the cattle with the same time serial number and obtain the state curve of the cattle in the whole cycle. The process flow chart of an average state time for several cattle can be found in \autoref{avecattle}.

\begin{figure}[ht]
    \centering
    \includegraphics[width=0.42\textwidth]{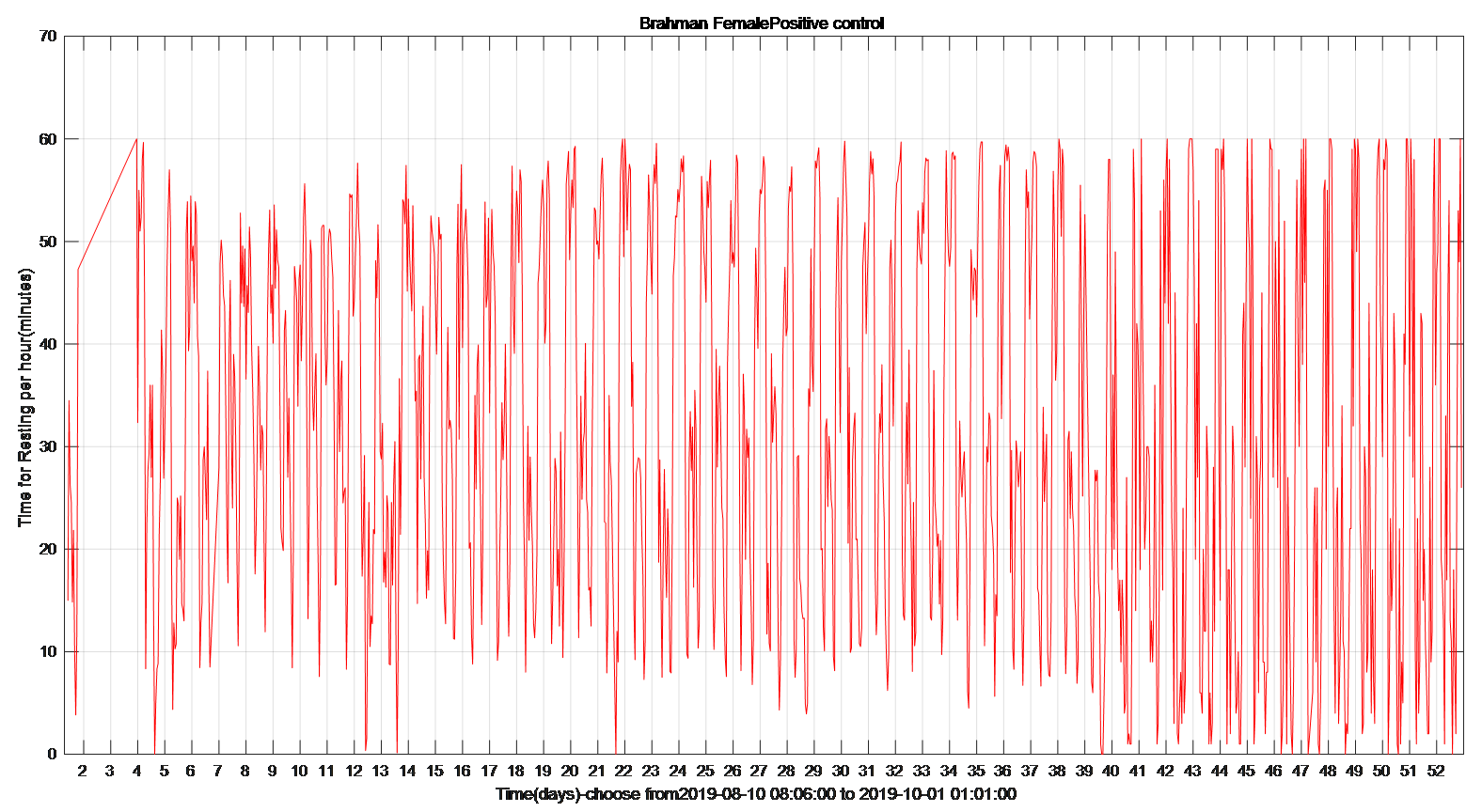}
    \caption{The resting time of Brahman Female with Positive control during the whole sample period.}
    \label{avecattleresults}
\end{figure}

The state diagram of cattle in the entire cycle can be obtained after the program has been executed. \autoref{avecattleresults} shows the calculated hourly rest time of the cattle in the whole cycle (Brahman Female with Positive Control). The number on the abscissa corresponds to the corresponding day, which includes all 24 hours. The ordinate represents the rest time corresponding to this hour in minutes.

\begin{figure}[ht]
    \centering
    \includegraphics[width=0.42\textwidth]{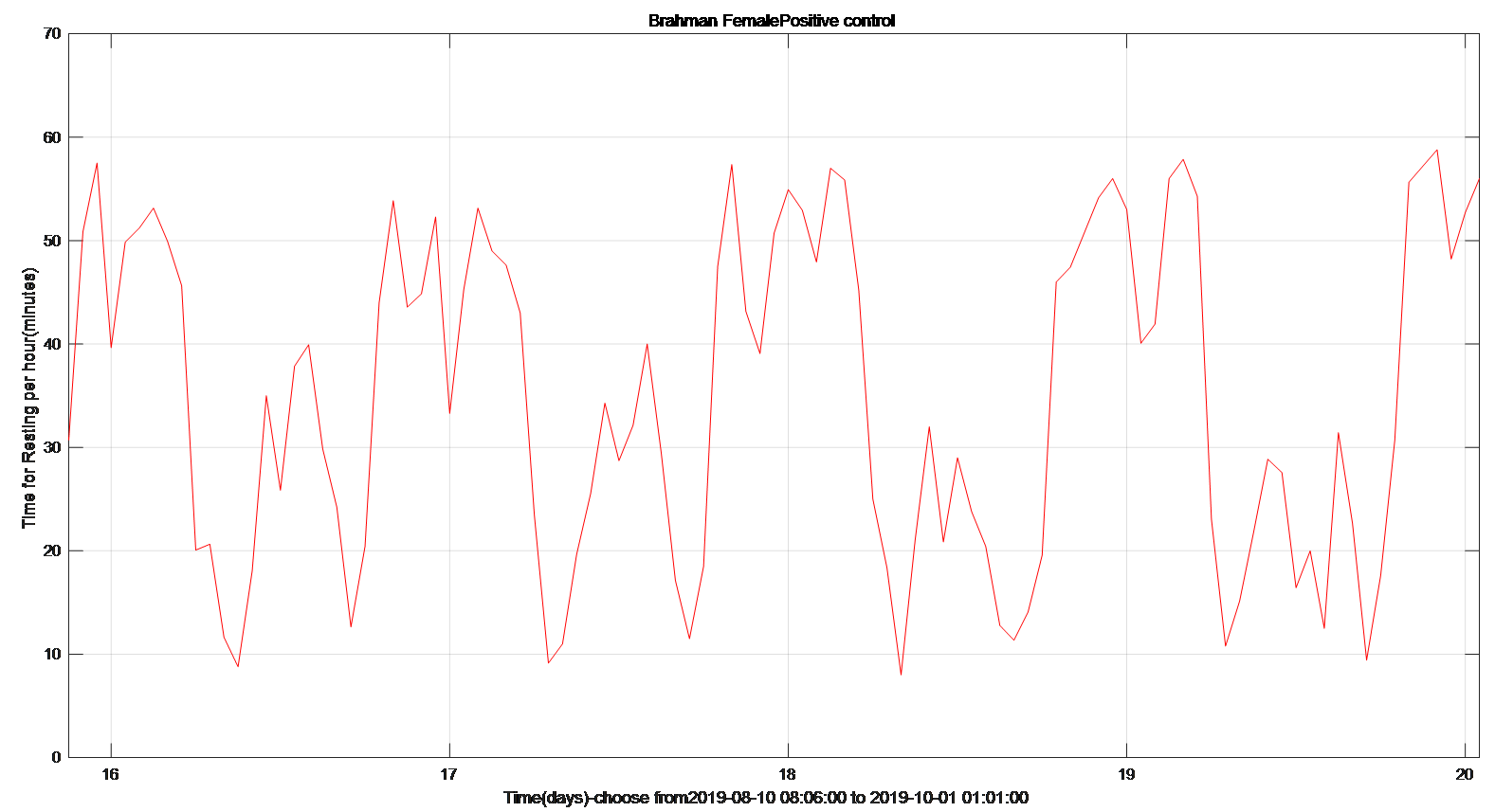}
    \caption{The enlarge figure of the resting time of Brahman Female with Positive control during the whole sample period.}
    \label{largeavecattleresults}
\end{figure}

\autoref{largeavecattleresults} is a detailed zoomed-in part of \autoref{avecattleresults} and located between days 16 and 20. It is obvious that the rest time of cattle varies periodically with a cycle of one day. The peaks of the daily rest time can be found in both early morning and late-night while the valleys can usually be identified at forenoon and afternoon hours.

\subsection{The average 24 hour state of cattle}
\begin{figure}[ht]
    \centering
    \includegraphics[width=0.42\textwidth]{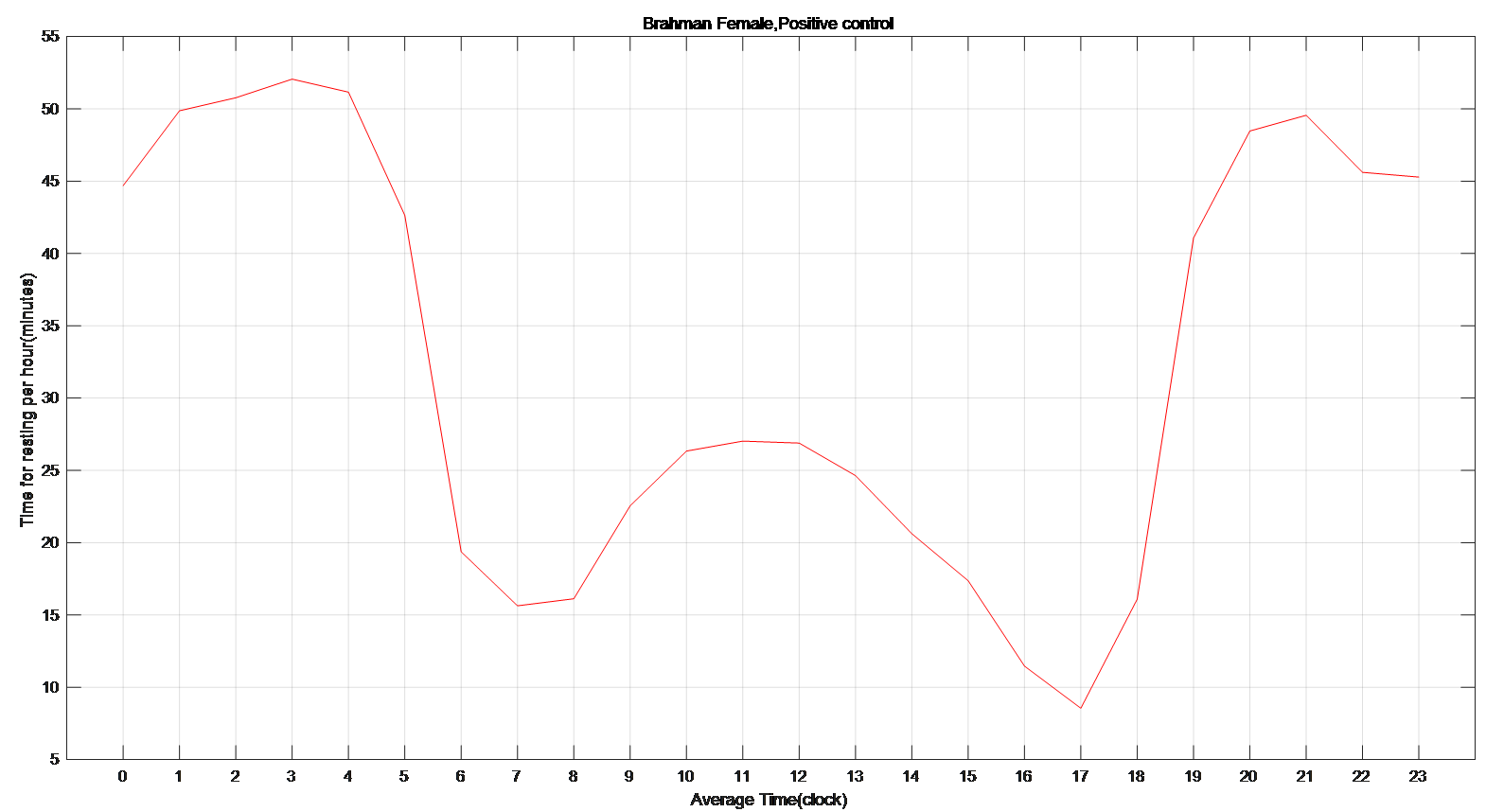}
    \caption{The average of a resting period for cattle(i.e., 24 hours a day)}
    \label{ave24hours}
\end{figure}

The averaged single rest cycle data result (which is 24 hours) of a single cattle is plotted in \autoref{ave24hours}. The entire sampling cycle is approximately 52 days as shown in \autoref{avecattleresults}. The abscissa refers to the o'clock, i.e., from 0:00 to 23:59, and the ordinate relates to the rest period in minutes at this hour (Brahman Female with Positive control). The average period's plot is flatter than a single period's plot. However, the trend and structure of these two are nearly identical, and a single cycle has more individual points and noises.

\subsubsection{Fitting curve for the average state period (24 hours)}
Curve fitting is commonly used to obtain the data relationship for such irregular curves. Typical fitting methods include minimum binomial fitting, exponential function fitting, power function fitting, and hyperbola fitting. Different fitting approaches are compared in this section to obtain the most ideal mathematical model\cite{48,49}.
\begin{table}[ht]
    \centering
    \caption{The results of different fitting methods.}
    \begin{tabu}{|X[0.35,c]|X[0.5,c]|X[0.35,c]|}
         \hline  
         Fitting method&The best number of items&Variance\\\hline
         Gaussian Fitting&8&3.0037\\\hline
         Sum of sine&8&20.1288\\\hline
         Polynomial&9&245.3264\\\hline
         Fourier&8&25.4590\\\hline
    \end{tabu}
    \label{differentfitting}
\end{table}

\begin{figure}[ht]
    \centering
    \includegraphics[width=0.45\textwidth]{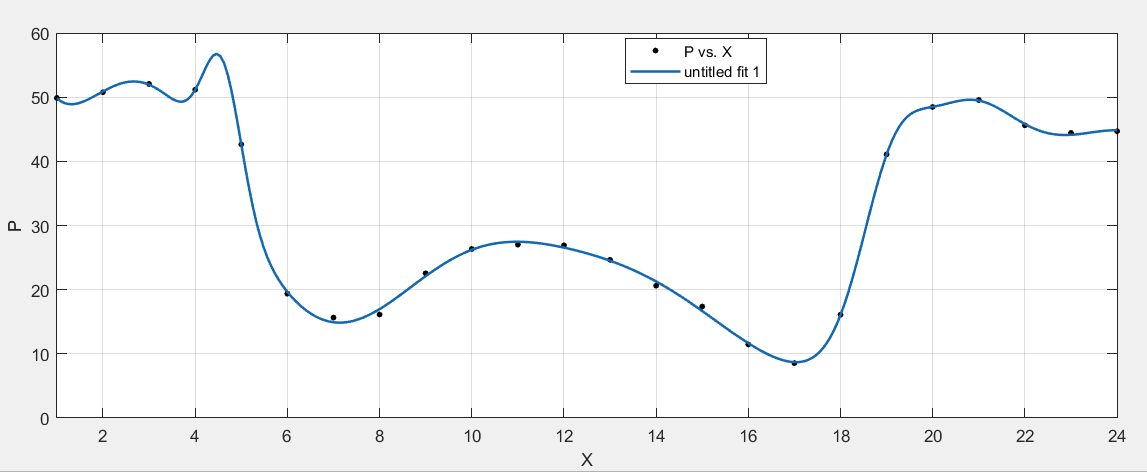}
    \caption{The Gaussian Fitting.}
    \label{gaussianfit}
\end{figure}

Using the MATLAB fitting toolbox and the Brahman Female with Positive control as an example, four fitting approaches are utilized to fit the 24-hour average rest duration of cattle: Gaussian fitting, Sum of Sine fitting, Polynomial fitting, and Fourier fitting. The independent variable is the time, and the dependent variable is the rest period of cattle corresponding to that time while fitting the curve. The relationship between the time and the associated rest time can be established, and the curve of the cattle's rest period throughout the day can be obtained. As indicated in \autoref{differentfitting}, Gaussian (item number 8) fitting is found to be the most accurate model among all candidates in terms of the fitting variance result. The fitted curve shape is depicted in \autoref{gaussianfit}. The formula of the fitting curve--Gauss eight-term formula is:
\begin{equation}
\begin{split}
    f(x) = &51.29e^{(-\frac{x-2.823}{2.957})^2} + 44.42e^{(-\frac{x-24.19}{3.936})^2} +\\ &1.378\times10^{14}e^{(-\frac{x+40.24}{7.546})^2} +
    19.29e^{(-\frac{x-13.55}{3.22})^2} +\\
    &16.18e^{(-\frac{x-19.06}{0.9367})^2} + 19.25e^{(-\frac{x-4.588}{0.5802})^2} + \\ &29.29e^{(-\frac{x-20.39}{1.802})^2} + 20.45e^{(-\frac{x-9.812}{2.834})^2}
    \label{fitequation}
\end{split}
\end{equation}

In \autoref{fitequation}, $x$ is the clock of a day, while $f(x)$ denotes the rest time within one hour of that clock. Regarding the low standard deviation and variance of this fitting result, this model is considered to be the proper candidate to describe the resting time of cattle in a day for Brahman Females with Positive control. The models for other breeds, genders, states, and combined therapies can be obtained in the same way.

\subsubsection{Compare the state time of cattle between different treatments}
\begin{table}[ht]
    \centering    
    \caption{The explanation of the state.}
    \begin{tabu}{X[1,l]|X[6,l]}
    \hline  
    State&Description\\
    \hline
    Rest&Standing still, lying, and transition between these 2 events. Allowed to move head and legs during standing if movement is only of short duration ($<$10 seconds), e.g. Head tossing associated with fly avoidance, stomping, briefly licking/sniffing self or environment but if movement of head/legs becomes dominant feature (e.g. sniffing/licking/chewing/pawing with leg or otherwise interacting with self/environment/other animal for more than 10 secs at a time) then no longer rest. While lying, allowed to do any kind of movement with head/neck/legs (e.g. tongue rolling). Only exception is paddling or otherwise struggling e.g. if secondary to being stuck under gate (which should then be classified as "undefined"). \\\hline
    Rumination&Rhythmic circular/side to side movements of jaw not associated with eating or medium activity, interrupted by brief ($<$5 seconds) pauses during time that bolus is swallowed, followed by continuation of rhythmic jaw movements. If ruminating, record as such regardless of body position or location in pen.\\\hline
    Panting (Heavy Breathing)&Respiratory rate$>$80, fast and shallow movement of thorax visible when looking animal from side, along with forward heaving movement of body while breathing. May or may not have open mouth, salivation, and/or extended tongue.\\\hline
    High Activity&Includes any combination of running, mounting, head-butting, repetitive head-weaving/tossing, leaping, buck-kicking, rearing and head tossing.\\\hline
    Eating&Muzzle/tongue physically contacts and manipulates feed, often but not always followed by visible chewing. May move from one location to another while eating, as long as break in eating doesn't last for more than 1 minute. More than 1 minute break in contact with feed and/or chewing-ends behavior. Searching for or otherwise manipulating in area of feed bunk or any other area in the pen in absence of feed is "bicycle".\\\hline
    Grazing&Eating (see above definition) growing grass and pasture, while either standing in place or moving at slow, even or uneven pace between patches.\\
    \hline 
    \end{tabu}
    \label{explanstate}
\end{table}
This part investigates and compares the performance of cattle in various states when given the combination therapy. To be more specific, the average 24-hour data from each state is compared to examine changes in the state of cattle due to various pain treatments. For Brahman Female cattle, just the dehorning procedure is performed, corresponding to four different pain treatments: Topical anesthetic, Meloxicam, Topical anaesthetic+Meloxicam, and Negative control. Resting, walking, panting (heavy breathing), grazing, and eating are the five state behaviours evaluated and compared in different pain therapies. Detailed descriptions for different cattle states, e.g., Rest, Rumination, Panting, High activity, eating and grazing  are explained in \autoref{explanstate}. 

\begin{figure}[ht]
    \centering
    \includegraphics[width=0.42\textwidth]{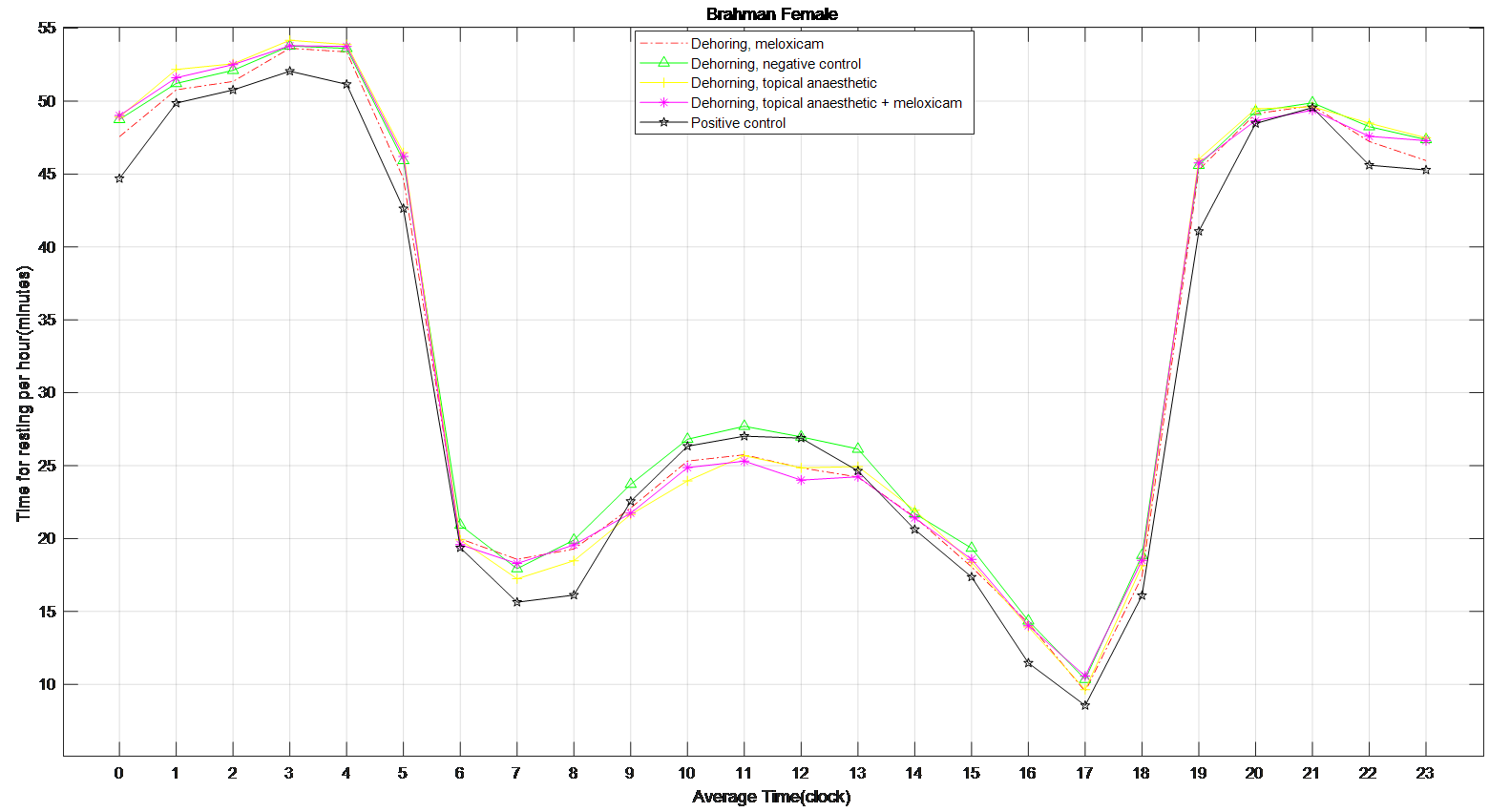}
    \caption{The resting dynamic of Brahman Female cattle doing different combined treatment in an average of 24 hours.}
    \label{rest}
\end{figure}

The Dehorning treatment is also employed for all the cattle, as shown in the dynamic comparison diagram of the cattles' rest in \autoref{rest}, and various pain therapies result in various states. The positive control, a comparison variable, is the cow that does not receive any combined therapy. When the cattle's horns are removed without pain relief, which is known as a negative control, they rest the longest. Except for the midday interval, cows treated with the other three anesthetic treatments rest slightly longer than those treated with a positive control and slightly shorter than those treated with a negative control. Nevertheless, in general, apparent similarity can still be found among their cases. This situation can be attributed to the rest state's subjective and environmental effect, as the difference in pain treatment cattle at rest is not significant.
\begin{figure}[ht]
    \centering
    \includegraphics[width=0.42\textwidth]{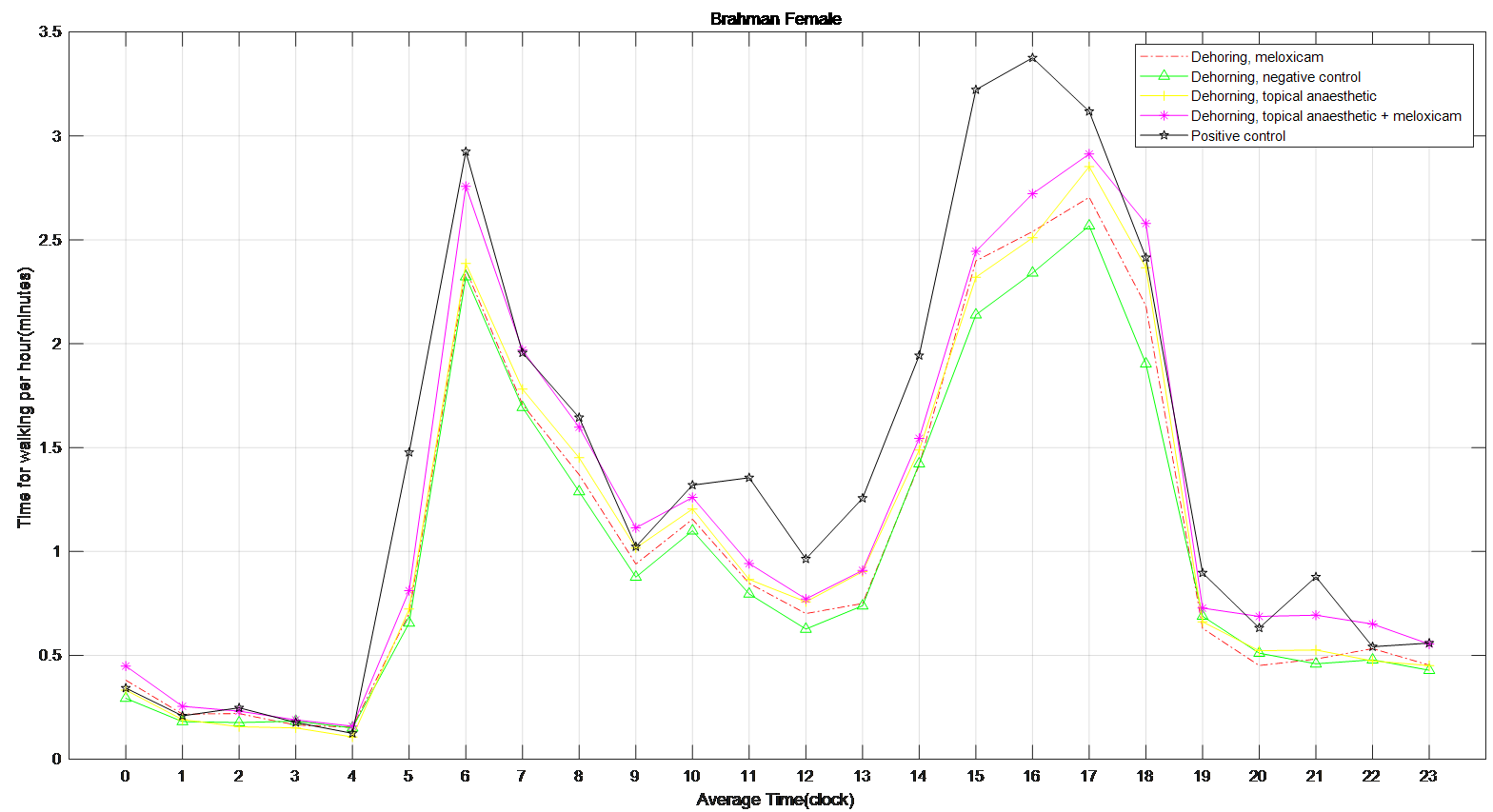}
    \caption{The walking dynamic of Brahman Female cattle doing different combined treatment in an average of 24 hours.}
    \label{walk}
\end{figure}

The cattle's walking dynamics vary more significantly after various pain treatments than the cow's resting dynamics. As shown in \autoref{walk}, after the dehorning treatment, every group walks for a shorter time than under the positive control. The cows provided with both topical anesthetic and Meloxicam are found to have the most prolonged walking duration, followed by those with topical anesthesia alone, and finally those with Meloxicam alone. However, cattle which is not anesthetized, such as Negative Control, have the least walking time. In summary, the cattle's walking time decreases with the dehorning but increases when the anesthetic is introduced. This phenomenon also reflects, to some extent, that cattle's pain condition directly ties to their walking time and that the administration of anesthetics affects cattle's walking dynamics.

\begin{figure}[ht]
    \centering
    \includegraphics[width=0.42\textwidth]{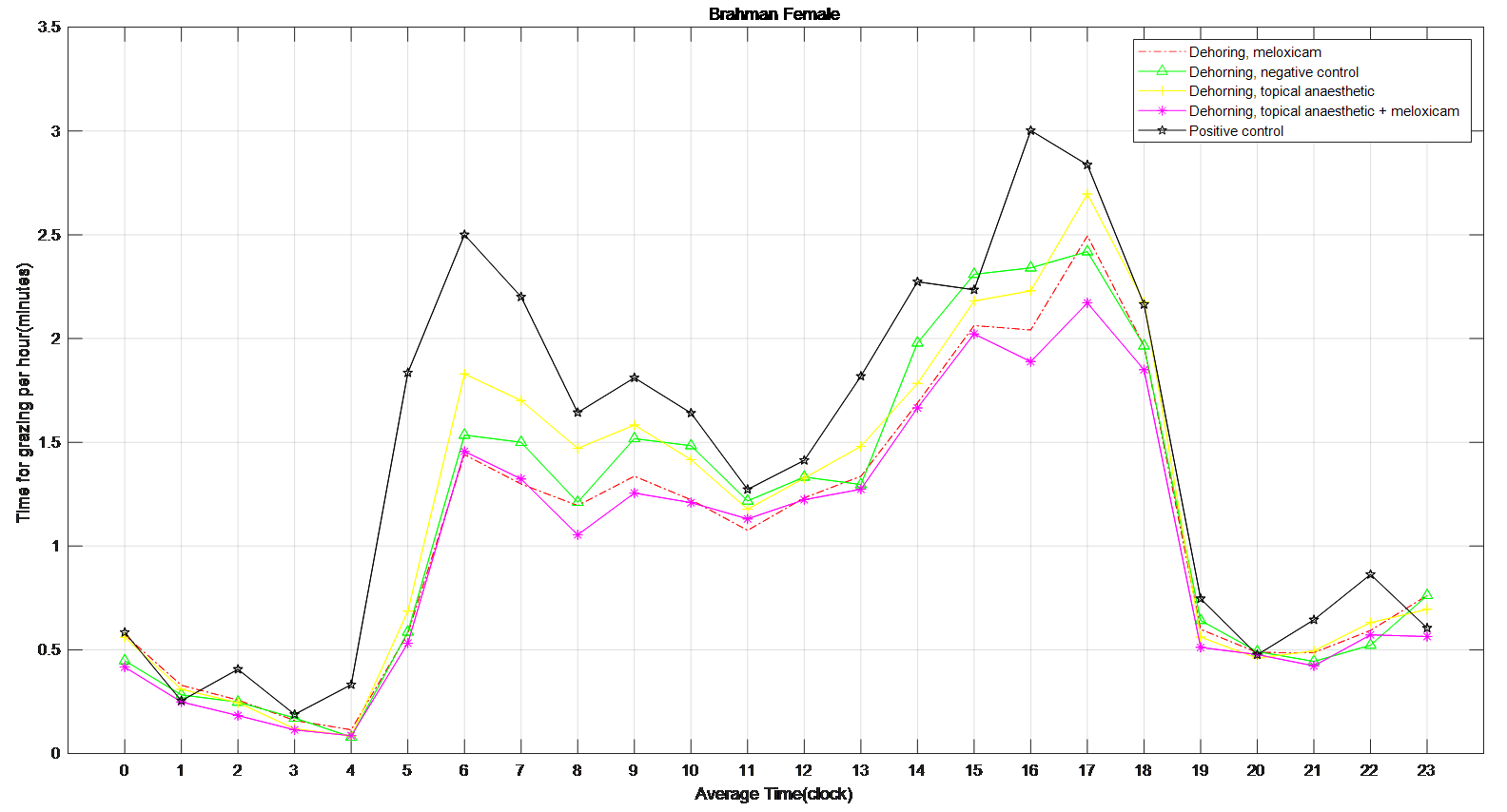}
    \caption{The grazing dynamic of Brahman Female cattle doing different combined treatment in an average of 24 hours.}
    \label{graz}
\end{figure}

The cattle grazing behaviours are also notably varied under different Pain treatments. The cattle treated with a positive control have the most prolonged grazing duration, followed by those treated with a negative control and those treated with a local anesthetic, as seen in \autoref{graz}. The cattle grazing takes much less time when cattle are given Meloxicam or both. Although the association between pain and grazing time cannot be demonstrated, it is clear that different anesthetic treatments have varying effects on grazing time and that the duration of the effects also varies correspondingly.

\begin{figure}[ht]
    \centering
    \includegraphics[width=0.42\textwidth]{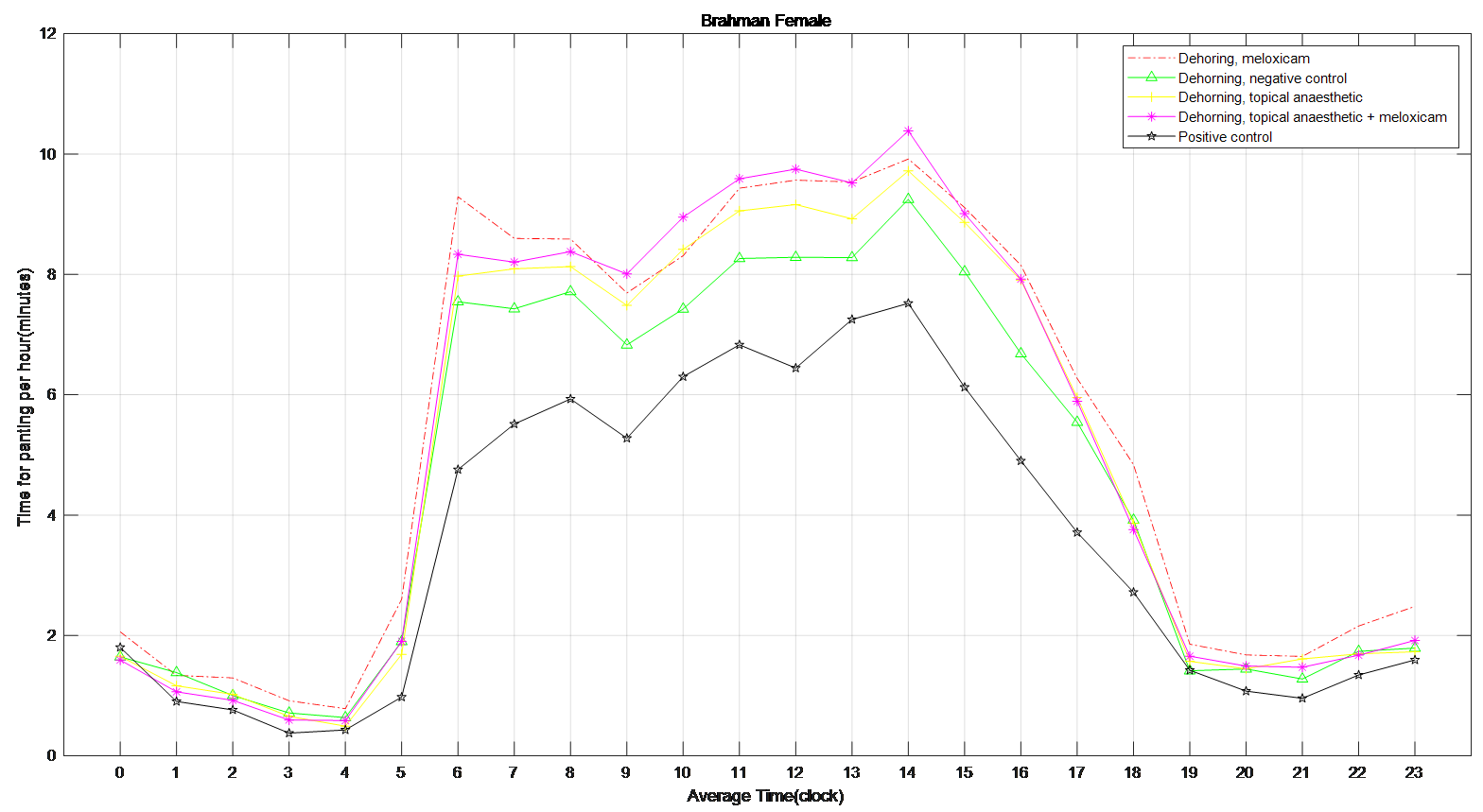}
    \caption{The panting dynamic of Brahman Female cattle doing different combined treatment in an average of 24 hours.}
    \label{pant}
\end{figure}

The most noticeable of these stages is the time to pant (defined as breathing heavily). The panting period is the longest when the calves receive Meloxicam or Meloxicam and topical anesthetic, as depicted in \autoref{pant}. There is no significant graphical difference between these two treatments. The panting time during topical anesthetic is the second factor, followed by the panting time under a hostile control. During the positive control, the amount of time spent on panting is the shortest. The cows only pant for less than 2 min/h in the early morning and late at night. However, most of the panting occur between 6 a.m. and 6 p.m., and the amount of panting change dramatically across pain treatments.

\begin{figure}[ht]
    \centering
    \includegraphics[width=0.42\textwidth]{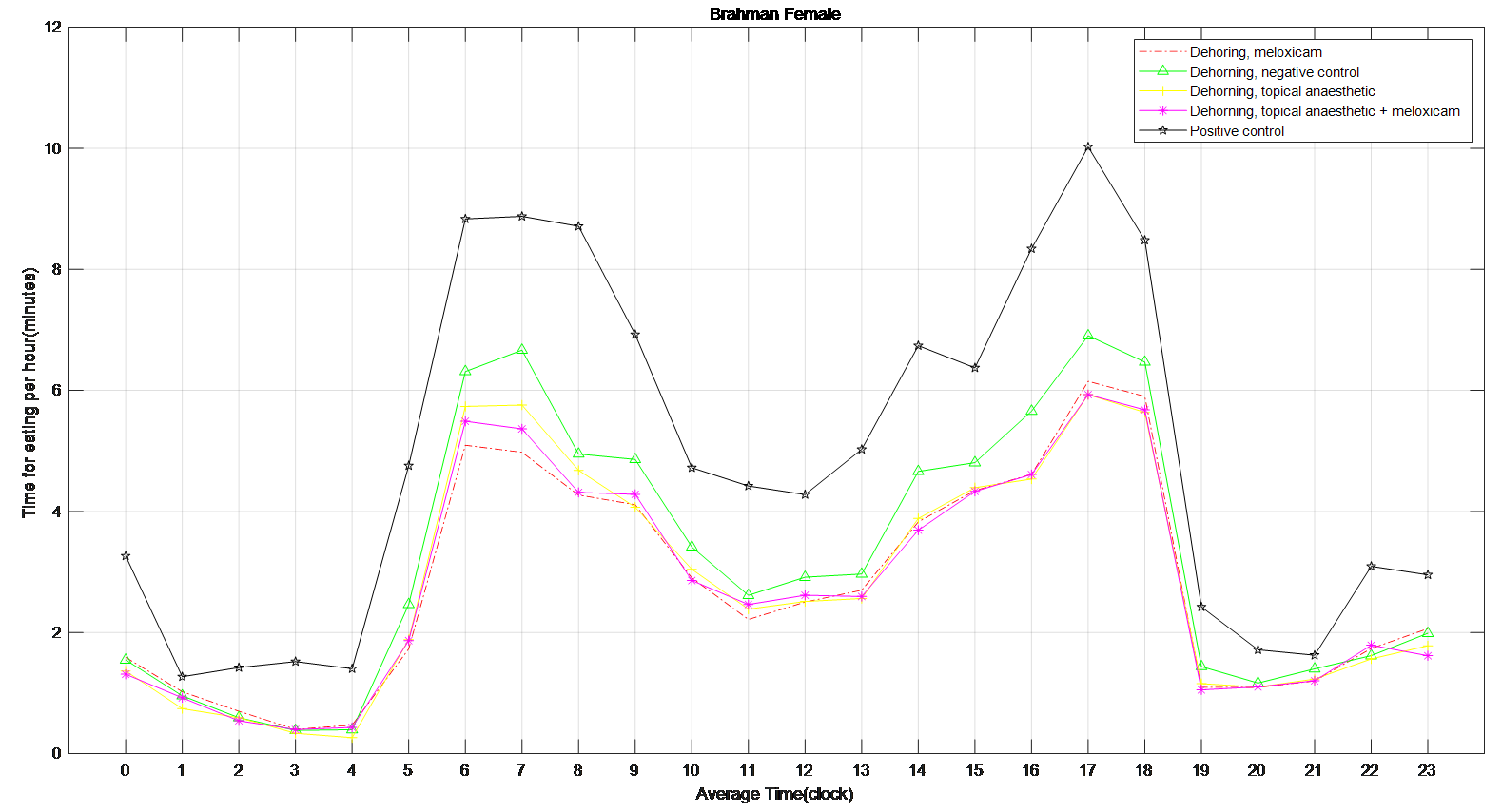}
    \caption{The eating dynamic of Brahman Female cattles doing different combined treatment in an average of 24 hours.}
    \label{eat}
\end{figure}

Cattle eat between 6 a.m. and 5 p.m., as shown in \autoref{eat}. The positive control group spends the most time eating, followed by the negative control group. The three with the anesthesia take less time than the negative control. As a result, the dynamic analysis at the eating time can distinguish between the cattle that do not receive any treatment (positive control) and the cattle that do not receive any treatment (negative control). However, the type of anesthesia can not be determined accurately.

\subsection{Noise reduction using low-pass FIR filter}
Throughout the sample period, the cattle's condition varies on daily basis. The plot of the entire activity cycle contains noise and outliers in \autoref{avecattleresults}. Therefore, denoising the sampled data is required.

FIR and IIR are two types of digital filters that are extensively employed. In theory, an IIR function's filtering effect is superior to that of an FIR function of the same order, but divergence can occur. The IIR digital filter has a high precision for amplitude-frequency characteristics, with a non-linear phase, it is suited for audio signals that are insensitive to phase information. FIR digital filters have lesser amplitude-frequency precision than IIR digital filters. However, the phase is linear, meaning the time difference between signals of various frequency components remains unaltered after going through the FIR filter. In addition, the calculation time delay is relatively tiny, it is suited for real-time signal processing\cite{50,51}. Because the state of the cattle is a time-series data, it is critical to ensure that the filtered phase remains constant. Therefore, in this work, we use the FIR low-pass filter for denoising.
\begin{figure}[ht]
    \centering
    \includegraphics[width=0.25\textwidth]{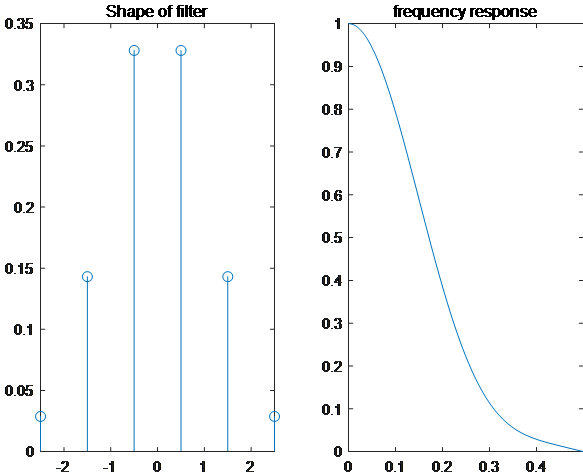}
    \caption{The shape of FIR filter and the frequency response.}
    \label{FIRshape}
\end{figure}

\begin{figure}[ht]
    \centering
    \includegraphics[width=0.42\textwidth]{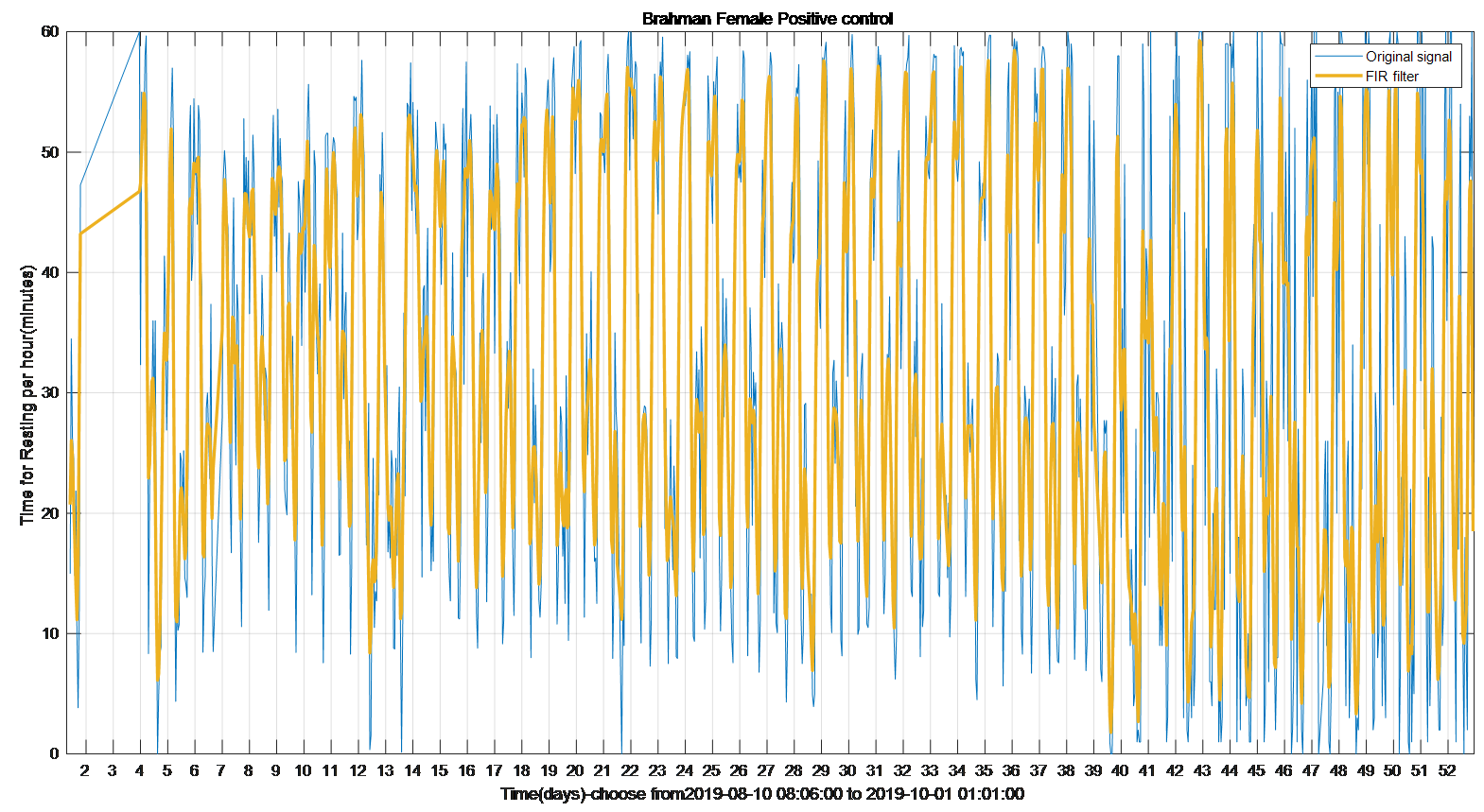}
    \caption{The resting time of cattle during the whole period after using the FIR filter.}
    \label{FIRfilter}
\end{figure}

Cattle monitoring data are sampled once every 60 seconds in this study, resulting in a sampling frequency of around 0.0167Hz. Noise frequency is more extensive than sampling frequency, so the signal between 0 and 0.0167Hz is kept while the signal above 0.0167Hz is eliminated. In \autoref{FIRshape}, the filter length is set to 5, and the filter's shape corresponds to its frequency. The filtered result is depicted in \autoref{FIRfilter}, which uses the resting time of Brahman Female's cow with Positive control as an example.

\begin{figure}[ht]
    \centering
    \includegraphics[width=0.42\textwidth]{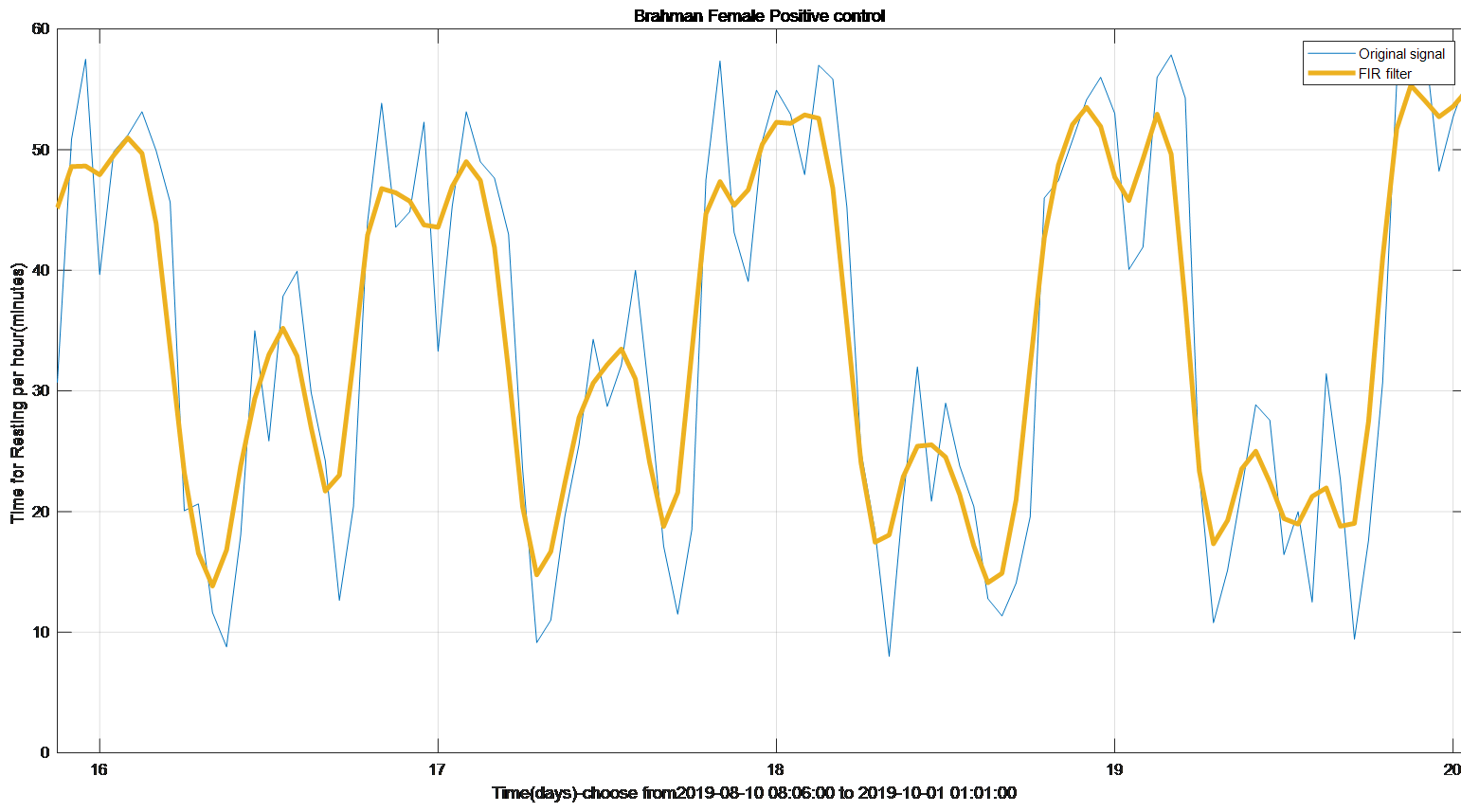}
    \caption{The large figure of the resting time of cattle during the whole period after using the FIR filter.}
    \label{largeFIRfilter}
\end{figure}

\autoref{largeFIRfilter} is a local detailed version of \autoref{FIRfilter}, focusing on the comparison of before using FIR filtering and after using FIR filtering from the 16th to the 20th day. Data performance is optimized after introduction of the FIR filter for smooth signal processing, and the data trend can be clearly identified.
\begin{figure}[ht]
    \centering
    \includegraphics[width=0.42\textwidth]{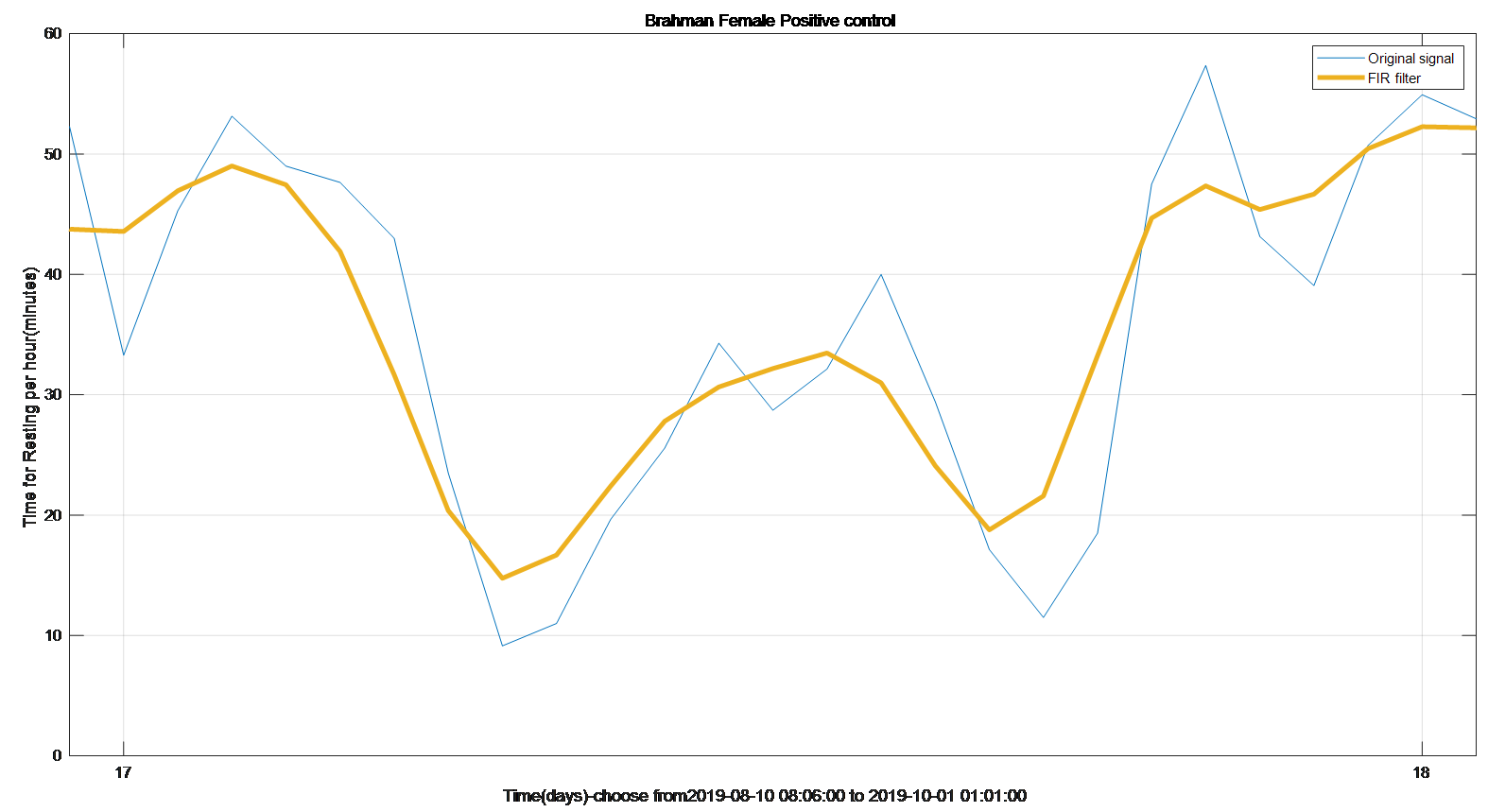}
    \caption{The single resting cycle after the FIR filter.}
    \label{oneFIRfilter}
\end{figure}
After going through the FIR filter, \autoref{oneFIRfilter} provides an image of a single rest period (one day, Day 17). In comparison to \autoref{ave24hours}, it exhibits the same trend, i.e., one day's rest time after filtering is nearly the same as one day's typical rest time. This feature demonstrates that the cattle's condition changes on a regular basis. It also indicates that the FIR filtered signal is effective and precise. The FIR filter effectively minimizes noise, and eliminates outliers, and gross inaccuracy. As a result, the signal filtered by the FIR filter can be used for subsequent modeling and prediction.

\section{Prediction based on LSTM model}\label{mainly}
In DL, LSTM network is a unique RNN model. Its unique structural design allows it to avoid long-term reliance. The default nature of LSTM is to remember information from a long time ago\cite{20,21,51,52}. In this section, we employ the LSTM model to forecast the status of cattle based on the above research content. To be more explicit, the structure and properties of LSTM and how to construct an LSTM model are first discussed. Second, using the LSTM model, the cattle status is modeled and forecast. Finally, the model is optimized in order to improve its accuracy.

\subsection{Build the LSTM model of the cattle state}
\begin{figure}[ht]
    \centering
    \includegraphics[width=0.38\textwidth]{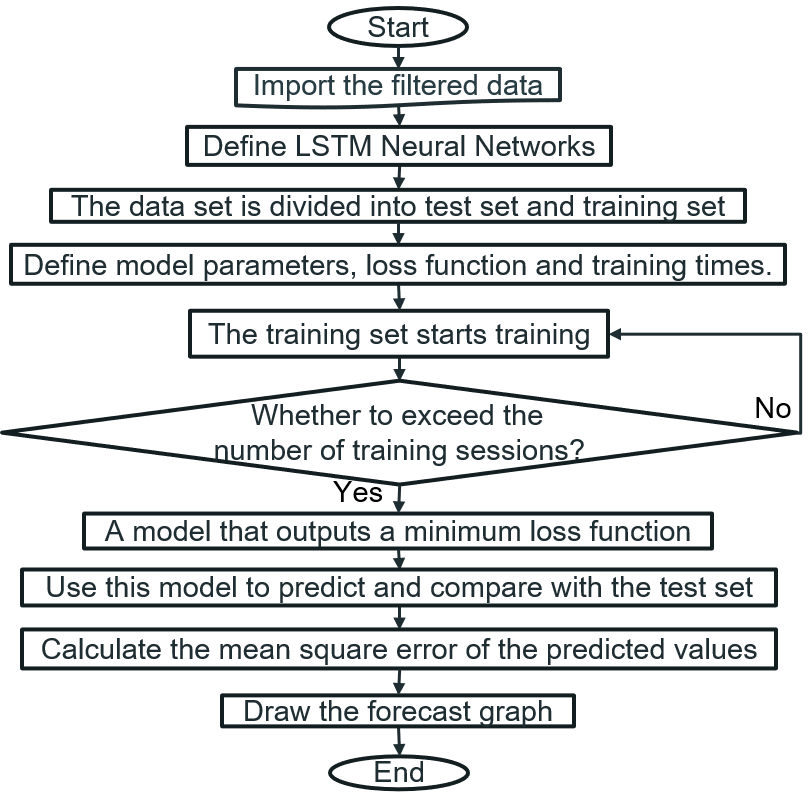}
    \caption{The code process of building the LSTM model.}
    \label{LSTMmodecode}
\end{figure}

The program flow chart for establishing the LSTM model is shown in \autoref{LSTMmodecode}. First, import the data previously filtered by the FIR filter, and divide it into a test set and a training set. Second, the LSTM model is created. Setting parameters: the number of input neurons, output neurons, hidden neurons, learning rate, batch size, epoch size (i.e., the number of training cycles) and the number of LSTM layers\cite{25}. The loss error is chosen as the mean square error, and the LSTM neural network is trained using the Adam optimisation technique\cite{25}. The cycle ends when the number of training times is reached, and the lowest loss error will be the output.

\subsection{Using the LSTM model to predict the state of cattle}
It is critical to determine the input, output, and time series before using the constructed LSTM model for cattle state prediction. The cattle's state must be presented as the output, and the number of the independent variable hours must be seen as a time series, according to the characteristics of the data sets. As a result, determining input variables is a challenging aspect of this approach. Because the output variable must be data with periodic changes, the input must be a known fixed periodic function. Time series as a fixed periodic function can be used as input. To be more specific, given that the state cycle of cattle is one day, it is appropriate to determine the input variable as the number of hours on the clock each day. The input and output variables, as well as the time series, for the resting time of Brahman Female's cattle with Positive control are as follows:
\\\textbf{Input:} The number of hours on the clock each day (24 hours).
\\\textbf{Output:} The resting time during this hour (e.g. The resting time at 7:00 means that the resting time during one hour from 7:00 to 7:59).
\\\textbf{Time series $t$:} The sequence number of this hour (e.g. 0:00am on the first day is the first hour, and $t$ is 1. So on, 0:00am on the second day is the 25th hour, and $t$ is 25).
\begin{itemize}
    \item \textbf{Training:}
          \\Both the input and output data are periodicity. The distinction is that the input in this cycle has a set value and trend, whereas the output in each cycle has a varied value. For example, the input is 0 at 0:00 am on Day 17th and 0:00 am on Day 24th, as shown by the two red lines in \autoref{LSTMinputoutput}, but the output is different. In other words, the same input might result in multiple outcomes regardless of time. Although the input is the same, the input's matching time series is not. As a result, when a single input correlates to numerous outputs in a time series, the LSTM model can successfully handle the problem.
          \begin{figure}[ht]
              \centering
              \includegraphics[width=0.42\textwidth]{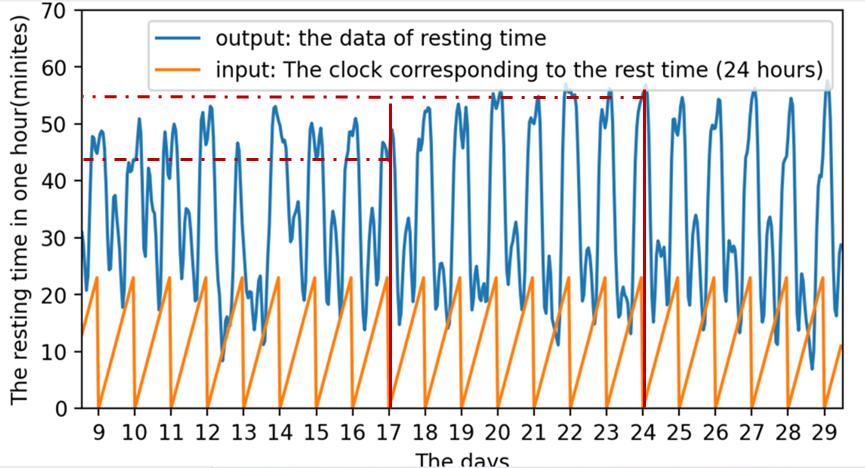}
              \caption{The input and output based on the LSTM model.}
              \label{LSTMinputoutput}
          \end{figure}
         
    \item \textbf{Testing and prediction:}
          \\90\% of the data is used for training, and 10\% for prediction and testing. For example, the input data sets for training are $input_{t_1}$ through $input_{t_{90}}$, while the data sets for testing are $input_{t_{91}}$ through $input_{t_{100}}$. The training outcomes are depicted in \autoref{LSTMpredic}.
          \begin{figure}[ht]
               \centering
               \includegraphics[width=0.2\textwidth]{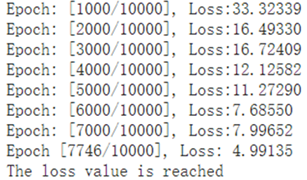}
               \caption{The training-loss during the LSTM model training.}
               \label{LSTMbadloss}
          \end{figure}
          \begin{figure}[ht]
              \centering
              \includegraphics[width=0.42\textwidth]{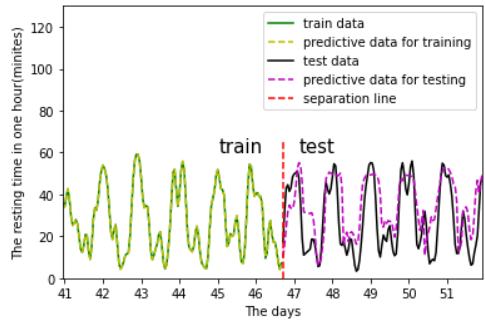}
              \caption{The predictive results after training and testing.}
              \label{LSTMpredic}
          \end{figure}
\end{itemize}

The predict and actual results are similar shown in \autoref{LSTMbadloss}. The training loss reduce during the training process, showing that model is converged and practical. However, the prediction results' error is relatively significant, which indicates further requirements of the parameter optimization in the model.

\subsection{Parameter optimization}
For optimization and comparison purpose, the number of hidden units, LSTM layers, the batch size, and the epoch size were all modified.
\\\textbf{Hidden units size:} 4, 8, 16, 32, 64, 128, 256.
\\\textbf{The number of LSTM layers:} 1, 2, 3, 4, 5, 6, 7.
\\\textbf{The batch size:} 3, 6, 12, 24, 48, 96.
\\\textbf{The epoch size:} 100, 500, 1000, 2000, 5000, 10000, 20000.
\begin{itemize}
    \item \textbf{Selection of the number of LSTM layers}
   \\The number of hidden units is 16, the batch size is 24, and the epoch size is 2000, all of which are randomly chosen. Only the number of layers in the LSTM is modified with the other parameters fixed: 1, 2, 3, 4, 5, 6, 7. The box diagram for the mean square deviation in the model learning process is shown in \autoref{Lstmlayers}.
    \begin{figure}[ht]
        \centering
        \includegraphics[width=0.43\textwidth]{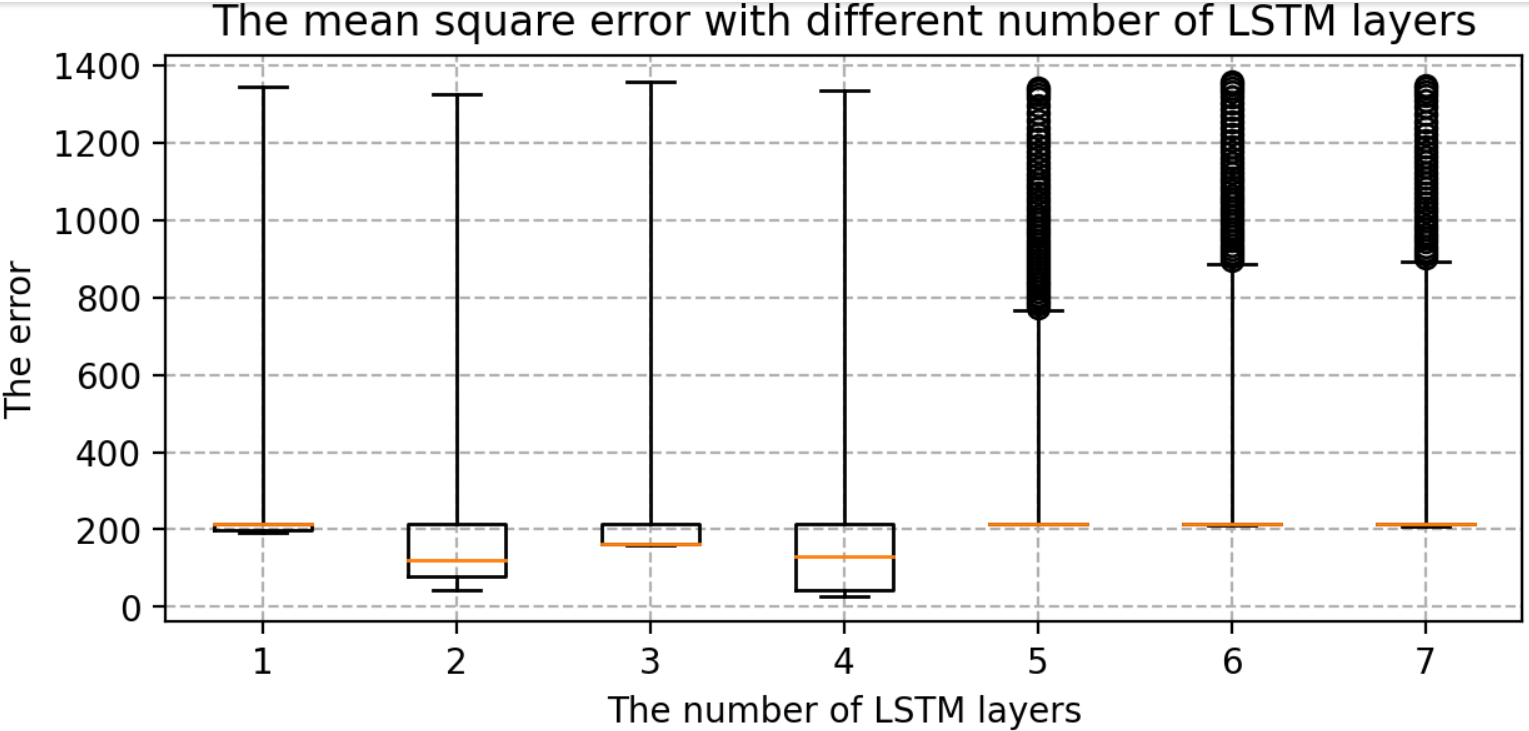}
        \caption{The mean square error with different number of LSTM layers.}
        \label{Lstmlayers}
    \end{figure}
    \\The top line and bottom line represent the edge's maximum and minimum values respectively. The upper quartile is represented by the box's upper edge, while the box's lower edge represents the lower quartile. The orange line represents the median. When comparing the seven box charts, increasing the number of layers has a minor impact on the mean square error of model training. However, in terms of model performance, using more LSTM layers, the running speed will be slower and it becomes more complex, and the result of the model operation is affected\cite{25}. As a result, two layers of LSTM are best for this model.
    \item \textbf{Selection of the hidden units size}
    \begin{figure}[ht]
        \centering
        \includegraphics[width=0.43\textwidth]{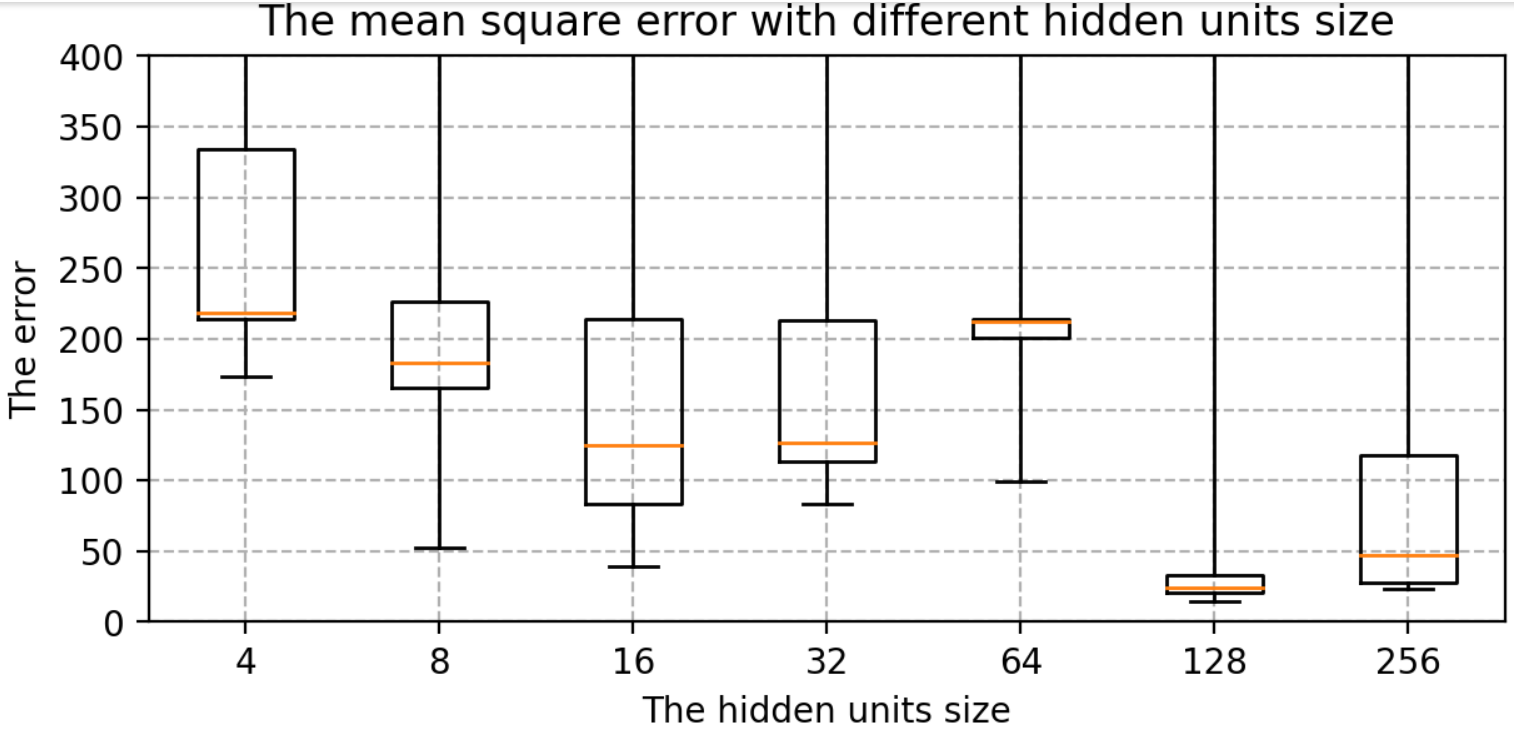}
        \caption{The mean square error with different hidden units size.}
        \label{hidden}
    \end{figure}
     \\To determine the size of the hidden units, we keep the batch size and epoch size unchanged and run the LSTM model with different hidden units size, i.e., 4, 8, 16, 32, 64, 128, 256. The box diagram of the mean square is shown in \autoref{hidden}. In terms of error size and ultimate training effect, the choice of 128 hidden units is the best for training the data, with the majority of the mean square error values falling below 25.
    
    \item \textbf{Selection of the epoch size}
     \\Select two layers of LSTM with 128 hidden units while keeping the rest of the settings the same: the batch size is 24, but the epoch size can be any of 100, 500, 1000, 2000, 5000, 10000, or 20000. \autoref{LSTMepoch} shows a box diagram for the mean square deviation in the model learning process.
    \begin{figure}[ht]
        \centering
        \includegraphics[width=0.43\textwidth]{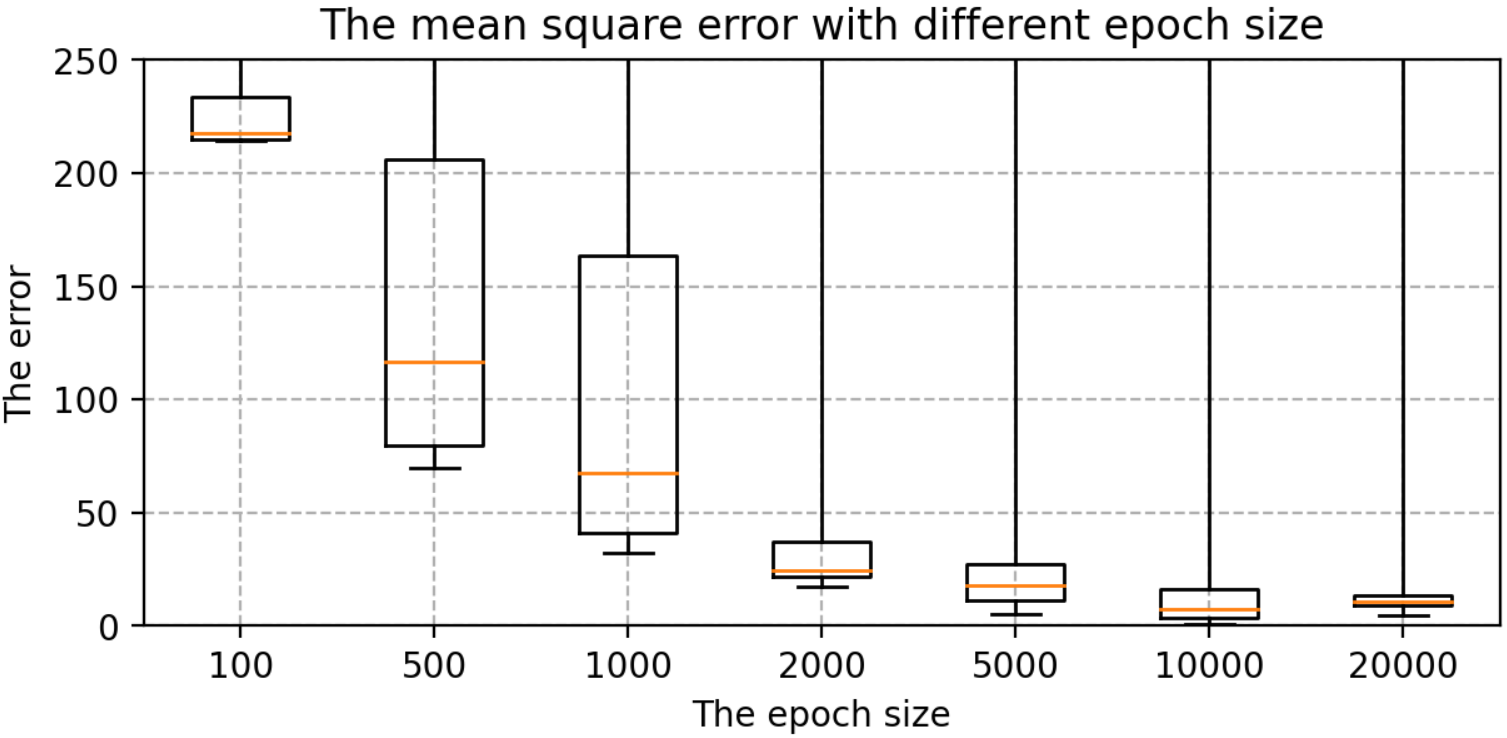}
        \caption{The mean square error with different epoch size.}
        \label{LSTMepoch}
    \end{figure}
    \\The epoch size is the number of times the learning algorithm works in the entire training data set. An epoch means that each sample in the training data set has the opportunity to update internal model parameters. In theory, the more training sessions there are, the better the fit and the lower the error. In practice, however, overfitting occurs when the epoch size exceeds a specific threshold, causing the training outcomes to deteriorate\cite{25}. The epoch size of 100, 500, 1000, 2000, 5000, 10000, and 20000 is chosen in \autoref{LSTMepoch}. The inaccuracy rapidly decreases and approaches zero as the epoch size increases from 100 to 10000. When the epoch size increases to 20,000, the error is still tiny, but it is greater than when the epoch size is 10,000, indicating an overfitting occurrence. Therefore, the model with a 10000 epoch size has the best effect.
    
    \item \textbf{Selection of the the batch size}
    \begin{figure}[ht]
        \centering
        \includegraphics[width=0.43\textwidth]{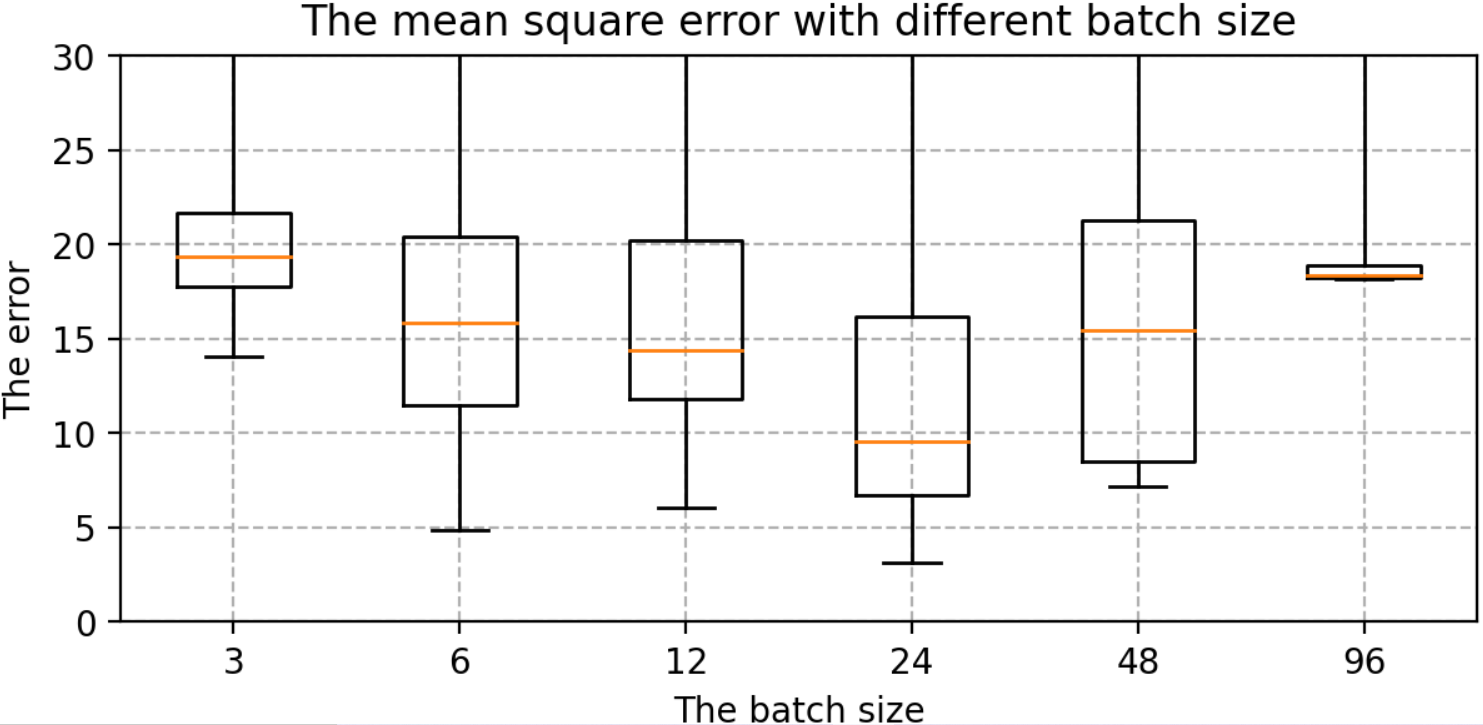}
        \caption{The mean square error with different batch size.}
        \label{LSTMbatch}
    \end{figure}
    \\The batch size, which can be 3, 6, 12, 24, 48, or 96, is altered when using two layers of LSTM with 128 hidden units and an epoch size of 10000. The box diagram is shown in \autoref{LSTMbatch}. The batch size refers to the number of samples fed into the model at once and divides the original data set into batch size data sets for independent training. This method helps to speed up training while also consuming less memory. To some extent, batch size training can help to prevent the problem of overfitting\cite{25}. As a result, when building the model, an acceptable batch size should be chosen. When the batch size is 24, the minimum value of the produced mean square deviation data set is the smallest in terms of minimum value and median.
\end{itemize}
\begin{figure}[ht]
    \centering
    \includegraphics[width=0.3\textwidth]{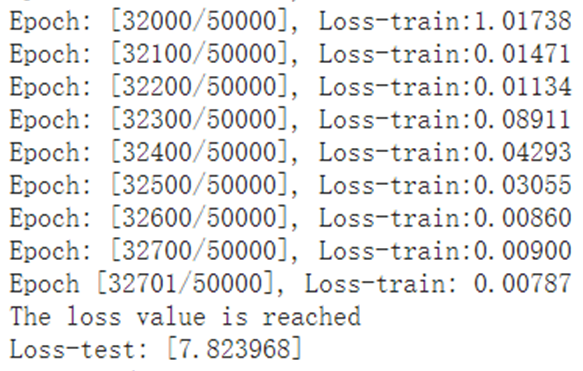}
    \caption{The loss value after parameter optimization.}
    \label{bestloss}
\end{figure}
\begin{figure}[ht]
    \centering
    \includegraphics[width=0.43\textwidth]{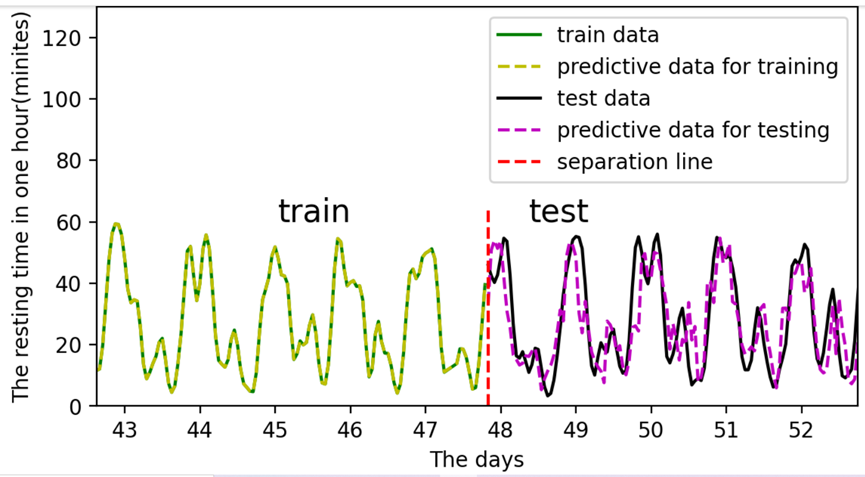}
    \caption{The training and prediction after optimizing model parameters.}
    \label{bestLSTMpredic}
\end{figure}

To sum up, the best parameters for the LSTM model are 128 hidden neurons, 24 batch size, 10000 epoch size, 2 LSTM layers. \autoref{bestloss} shows the loss value after optimizing parameters, while \autoref{bestLSTMpredic} shows the training and prediction outcomes after optimizing model parameters. The LSTM model has a good prediction on the resting state of cattle, which largely adheres to the periodic changes in cattle state and has a modest error. Loss-train: 0.00787; Loss-test: 7.823968.

\section{Results and Analysis}\label{Result}
In this section, the digital twin model of cattle is analyzed and evaluated to determine its reliability and accuracy. In addition, this section conducts a behavioural evaluation on the status of cattle. It evaluates the pain conditions of cattle based on the results produced from the previous sections.

\subsection{Applicability of the model} 
Figs.\ref{othersex}-\ref{otherstate} depicts the LSTM model's training and prediction on different sexes, breeds, combined treatments, and states, respectively.

The trend of the results predicted by this LSTM model are nearly identical to the actual data. The model for Brahman males performs relatively poorly, which can be attributed to their relatively random rest state, poor cycle regularity, and other external environmental factors. It is possible that increasing the size of the data collection may result in improved predictions. Overall, the LSTM-based model for cattle states cycle is accurate and effective, and it can accurately predict the dynamic trend of the next cattle state cycle.
\begin{itemize}
    \item \textbf{The LSTM model is applied to the other gender of cattle.}
    \begin{figure}[ht]
        \centering
        \includegraphics[width=0.43\textwidth]{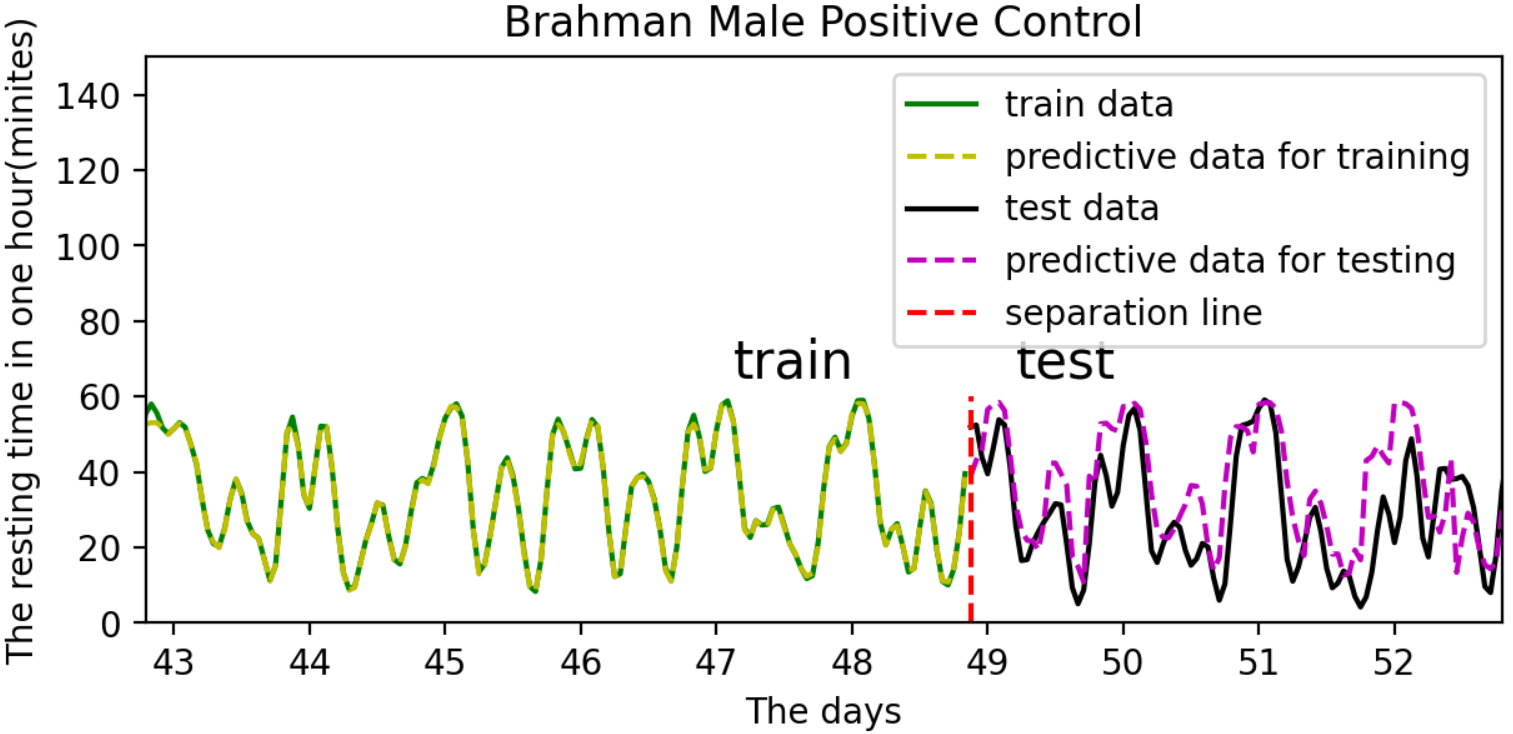}
        \caption{Prediction of rest dynamics during the whole cycle of Brahman Male cattle with Positive control.}
        \label{othersex}
    \end{figure}
    \item \textbf{The LSTM model is applied to the other breed of cattle.}
    \begin{figure}[ht]
        \centering
        \includegraphics[width=0.43\textwidth]{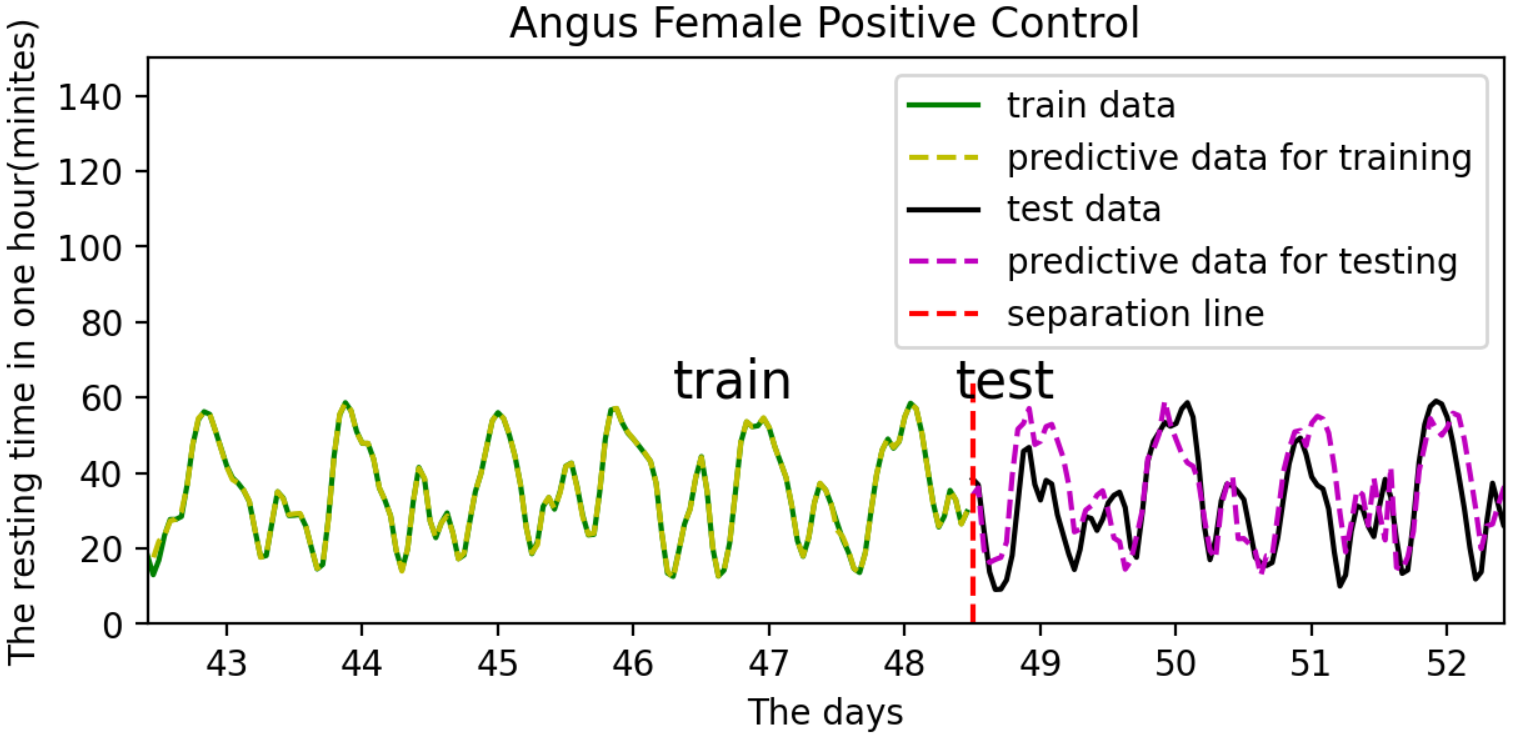}
        \caption{Prediction of rest dynamics during the whole cycle of Angus Male cattle with Positive control.}
        \label{otherbreed}
    \end{figure}
    \item \textbf{The LSTM model is applied to the other combined treatment of cattle.}
    \begin{figure}[ht]
        \centering
        \includegraphics[width=0.43\textwidth]{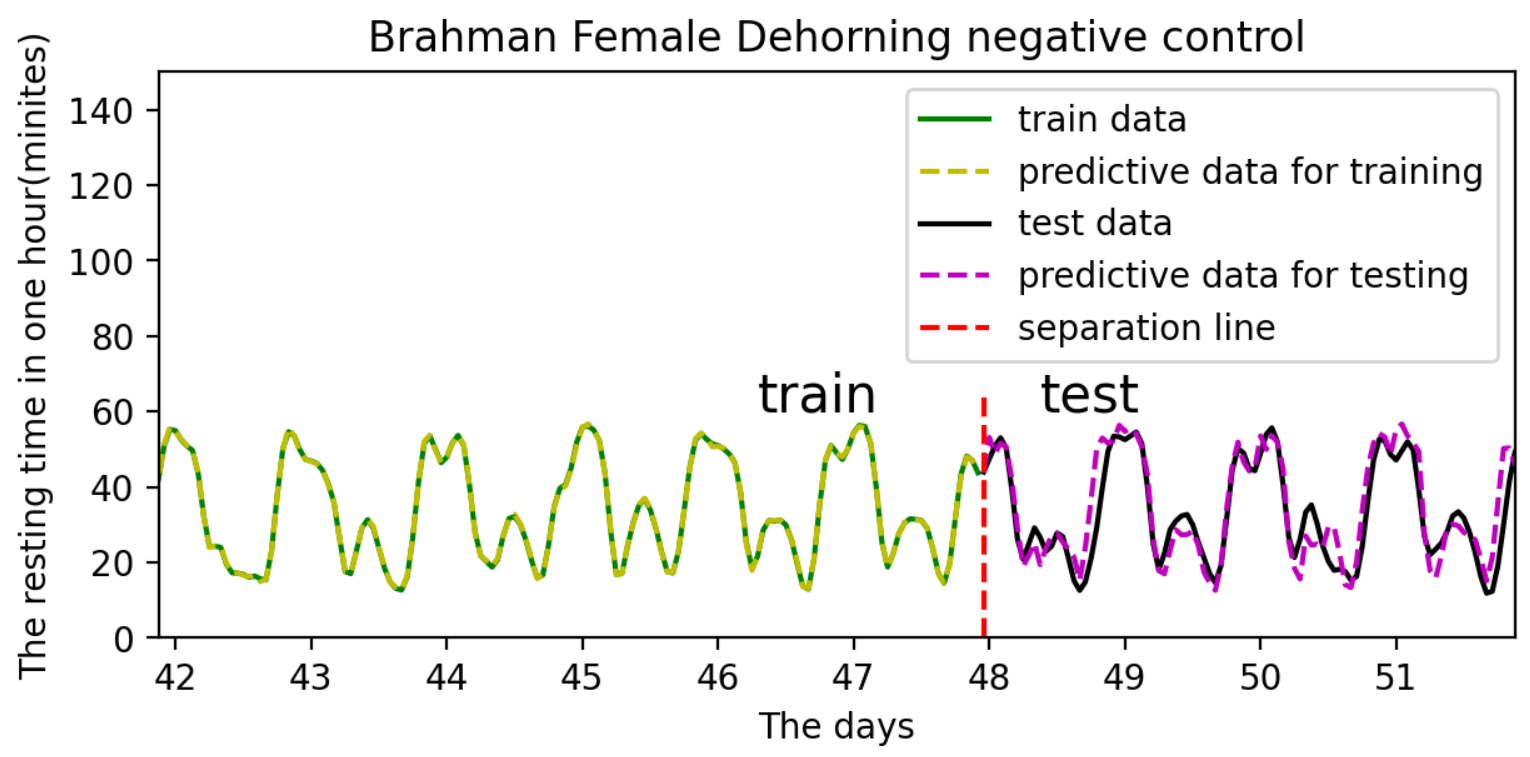}
        \caption{Prediction of rest dynamics during the whole cycle of Brahman Female cattle with dehorning treatment plus negative control.}
        \label{othertreat}
    \end{figure}
    \item \textbf{The LSTM model is applied to the other state of cattle.}
    \begin{figure}[ht]
        \centering
        \includegraphics[width=0.43\textwidth]{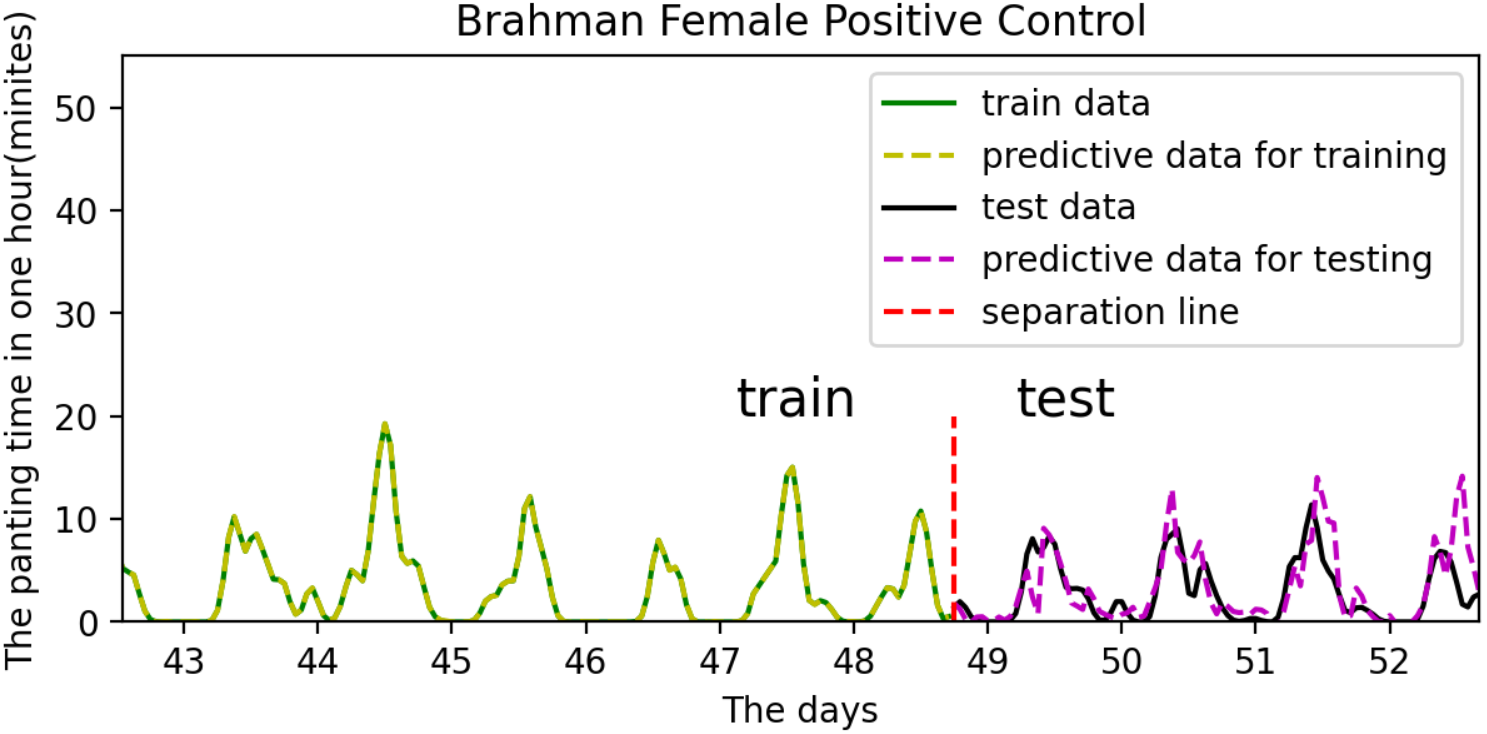}
        \caption{Prediction of pant dynamics during the whole cycle of Brahman Female cattle with Positive control.}
        \label{otherstate}
    \end{figure}
\end{itemize}
 
\subsection{The relationship between cattle's states and pain treatment}
This work primarily involves three different pain treatments for pain management: topical anesthetic, meloxicam, and a combination of the two. In animal practice, topical anesthetics are the most widely utilized pre-emptive analgesics\cite{53}. Without causing loss of consciousness, these chemicals cause a reversible loss of sensation in the local area. Local anesthetics enter nerve cells and block open sodium channels, inhibiting nerve impulse production and transmission. However, topical anesthetic must work in an alkaline environment, and the quality of local anesthesia in diseased, ischemic, or damaged tissue is frequently poor because the more acidic environment inhibits the medication from being separated\cite{41,53}. Topical anesthetics typically take 2 to 5 minutes to take effect, and studies have shown that they reduce plasma cortisol concentrations following castration, but not the mean area under the cortisol effect curve\cite{41}. This issue shows that using topical anesthetics alone can help reduce the immediate pain associated with castration\cite{41}. Meloxicam, an oxicam-class NSAID, has been licensed for the adjuvant treatment of acute respiratory illnesses in the European Union. Meloxicam muscle injection and horn nerve block lowered serum cortisol response for longer than cattle with a topical anesthetic before cauterization. This phenomenon means that the use of meloxicam before horn removal is effective in reducing pain in cattle. However, meloxicam alone is not effective in reducing castration-related acute distress. As a result, using a combination of topical anesthetic and meloxicam to reduce discomfort during castration or dehorning is beneficial\cite{41}.

\section{Conclusion}\label{sec:Conclusion}
\subsection{Achievements at the current stage}
The construction of a smart digital twin model of the state of cattle is primarily achieved in this work. It is primarily built on a farm IoT system to collect the state data of cattle under various combined treatments, with data cleaning and calculating. The average data of 24 hours are fitted, and the data of the whole sampling period are de-noised. In addition, a deep learning-based LSTM model for cattle state dynamics is developed using the data after noise reduction, and the model can predict the state change of cattle in the next cycle. The model's accuracy and effectiveness are demonstrated when the prediction results are compared to the actual results.

Furthermore, the association between various behaviour patterns and bovine pain, the relationship between pain degree and behaviour pattern of cattle under various pain treatments are investigated. It is fund that the cattle in pain spent less time walking and ate less food, according to the final data. In cattle undergoing castrating or dehorning, combining topical anesthetic with meloxicam can be observed to successfully reduce the cattle's pain.

\subsection{Future outlook}
Because pain is regarded as an emotion with a high subjective aspect, and the cattle behaviour is significantly impacted by external influences, existing data and technology do not quantify pain in a fully effective fashion. As a result, simply connecting the state of a cow with pain may not be accurate or helpful enough. Quantifying pain and measuring the intensity of discomfort in proportion to the duration of each behavioural state is a critical challenge that will need to be addressed in the future. Furthermore, encapsulating the entire research into another system is a critical step towards commercializing digital twins in the future.


\end{document}